\documentclass[runningheads]{llncs}
\usepackage{graphicx}
\usepackage{tikz}
\usepackage{comment}
\usepackage{graphicx}
\usepackage{array}
\usepackage{booktabs}
\usepackage{amsmath,amssymb} % define this before the line numbering.
\usepackage{color}

\usepackage{multirow}
\usepackage{amsfonts}
\usepackage{subfigure}
\usepackage{newfloat}
\usepackage{listings}

\usepackage[accsupp]{axessibility}  % Improves PDF readability for those with disabilities.

\begin{document}
% \linenumbers
\pagestyle{headings}
\mainmatter
\def\ECCVSubNumber{6262}  % Insert your submission number here

\title{Multimodal Adaptive Distillation for Leveraging Unimodal Encoders for Vision-Language Tasks}
%\title{Multimodal Adaptive Distillation for Leveraging Unimodal and Multimodal Pre-Training Models}
%\title{Leveraging Pretrained Unimodal Attention-based Encoders for Vision-Language Tasks with Only Finetuning}
%\title{Fusing large-scale Pretrained Vision, Text and Vision-Language Models with Finetuning} % Replace with your title

%\begin{comment}
% \titlerunning{ECCV-22 submission ID \ECCVSubNumber}
% \authorrunning{ECCV-22 submission ID \ECCVSubNumber}
% \author{Anonymous ECCV submission}
% \institute{Paper ID \ECCVSubNumber}
%\end{comment}
%******************

% \begin{comment}


\titlerunning{Multimodal Adaptive Distillation for Leveraging Unimodal Encoders for Vision-Language Tasks}

\author{Zhecan Wang\inst{1}\thanks{Equal Contribution.}
  \and Noel Codella\inst{2}$^*$
  \and Yen-Chun Chen\inst{2}
  \and Luowei Zhou\inst{2}
  \and Jianwei Yang\inst{2}
  \and Xiyang Dai\inst{2}
  \and Bin Xiao\inst{2}
  \and Haoxuan You\inst{1}
  \and Kai-wei Chang\inst{3}
  \and Shih-Fu Chang\inst{1}
  \and Lu Yuan\inst{2}}

  \authorrunning{Wang et al.}

\institute{Columbia University \and
Microsoft Research \and University of California, Los Angeles}

% \thanks{Equal Contribution. Correspondence to: Zhecan Wang $\langle$olinzhecanwang@gmail.com$\rangle$, Noel Codella $\langle$ncodella@microsoft.com$\rangle$.}

% \footnotetext[1]{Columbia University}
% \footnotetext[2]{Microsoft Research}
% \end{comment}
%******************
\maketitle

\begin{abstract}

Abstract. Cross-modal encoders for vision-language (VL) tasks are often pretrained with carefully curated vision-language datasets. While these datasets reach an order of 10 million samples, the labor cost is prohibitive to scale further. Conversely, unimodal encoders are pretrained with simpler annotations that are less cost-prohibitive, achieving scales of hundreds of millions to billions. As a result, unimodal encoders have achieved state-of-art (SOTA) on many downstream tasks. However, challenges remain when applying to VL tasks. The pretraining data is not optimal for cross-modal architectures and requires heavy computational resources. In addition, unimodal architectures lack cross-modal interactions that have demonstrated significant benefits for VL tasks. Therefore, how to best leverage pretrained unimodal encoders for VL tasks is still an area of active research. In this work, we propose a method to leverage unimodal vision and text encoders for VL tasks that augment existing VL approaches while conserving computational complexity. Specifically, we propose Multimodal Adaptive Distillation (MAD), which adaptively distills useful knowledge from pretrained encoders to cross-modal VL encoders. Second, to better capture nuanced impacts on VL task performance, we introduce an evaluation protocol that includes Visual Commonsense Reasoning (VCR), Visual Entailment (SNLI-VE), and Visual Question Answering (VQA), across a variety of data constraints and conditions of domain shift. Experiments demonstrate that MAD leads to consistent gains in the low-shot, domain-shifted, and fully-supervised conditions on VCR, SNLI-VE, and VQA, achieving SOTA performance on VCR compared to other single models pretrained with image-text data. Finally, MAD outperforms concurrent works utilizing pretrained vision encoder from CLIP. Code will be made available.

\end{abstract}

\section{Introduction}

\begin{figure}[t!]
\resizebox{12.8cm}{!}{%
\centering
\includegraphics[width=1\linewidth]{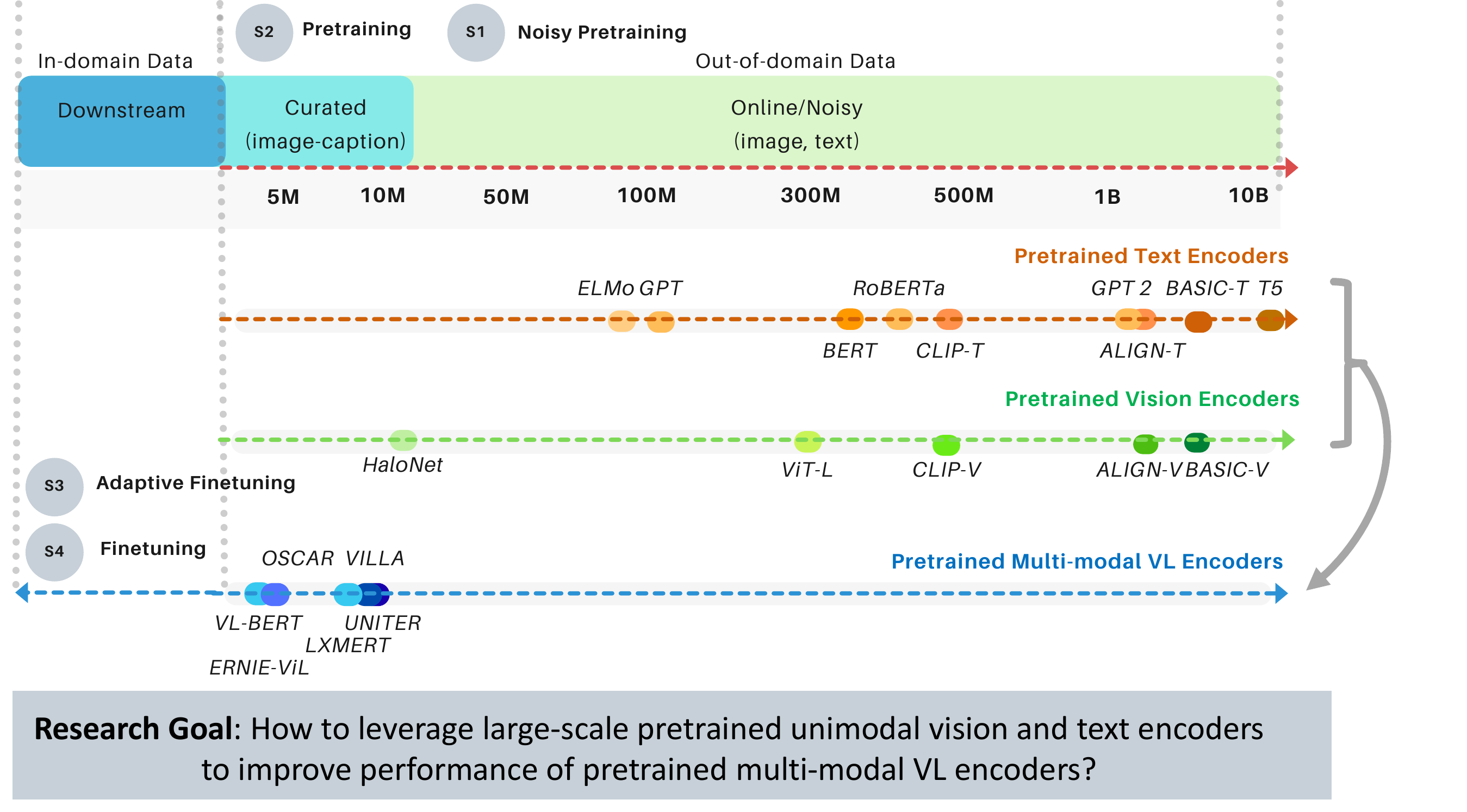}
}
\caption{Comparison of data size between different pretraining frameworks. It is noticeable that the data size of pretraining unimodal encoder models is generally much larger than Vision-Language models.}
% \vspace{-5mm}
\label{fig:pretraindatasizecompare}
% \vspace{-2mm}
\end{figure}

Visual Commonsense Reasoning (VCR) \cite{zellers2019vcr}, Stanford Natural Language Inference Visual Entailment (SNLI-VE) \cite{xie2019visual}, and Visual Question Answering (VQA) \cite{VQA}, are all VL tasks requiring solutions that effectively bridge the two modalities, with mechanisms of logic and prior knowledge serving as the link between them. For example, in VCR, logical reasoning capability and common knowledge are useful in answering questions about images, especially when the visual information is ambiguous, rendering the task of connecting questions with their correct answers difficult. Considering this, most recent works that have achieved SOTA tend to capture the logic and prior knowledge with various cross-modal transformer architectures that jointly model both modalities in a unified architecture ~\cite{chen2020uniter,Su2020VL-BERT,gan2020large,huang2021seeing,huang2020pixel,lxmert,wang2021sgeitl}. These models are typically pretrained on large images-caption dataset \cite{sharma2018conceptual,lin2014microsoft}. These curated datasets are all collected carefully by human annotators to ensure the text data consists of visually descriptive and syntactically correct sentences. They typically do not exceed about 10M data points in size.

% Existing top-performing Vision-Language (VL) models on VL tasks especially the highly semantic ones tend to utilize large image caption datasets \textit{e.g.} [COCO Captions, Conceptual Captions, xxxx] for large-scale pretraining. The

By contrast, obtaining training data for unimodal models requires significantly less effort and manual labor, although the data tends to have large domain gaps against downstream VL tasks. For example, attention-based text models like BERT \cite{devlin2018bert}, GPT \cite{radford2019language}, RoBERTa \cite{roberta}, BEiT \cite{bao2021beit}, \textit{etc.} can conveniently leverage self-supervision or weak-supervision to conduct unimodal pretraining on large text corpus \cite{sharma2018conceptual,lin2014microsoft}. Vision models like ViT \cite{dosovitskiy2020image} can also be obtained via unimodal supervised pretraining on large image datasets such as JFT-300M \cite{sun2017revisiting}. Furthermore, another line of research has been examining pretraining with paired noisy image-text data (crawled from the web) which is also easier to collect than image-caption pairs. This pretraining is typically combined with a contrastive learning objective to align both modalities into a shared embedding space. In these frameworks, vision, and text streams are all modality-specific encoders and cross-modal fusion layers do not exist but shallow connections via cosine distance. Thus different from the conventional VL frameworks and similar to unimodal pretraining, those contrastive pretraining can also produce pretrained unimodal encoder models like CLIP-V, CLIP-T, \textit{etc.} These works extend pretraining datasets to a more massive scale, such as CLIP \cite{clip} with 400 million, ALIGN \cite{jia2021scaling} with 10 billion and BASIC \cite{basic} with 100 billion (the largest so far).

These pretrained unimodal encoder models are generally pretrained with larger data than pretrained VL models, as shown in Fig. \ref{fig:pretraindatasizecompare}\footnote{The unit is sample data point. One data point can represent an image in visual pretraining and phrases, sentences in text pretraining.}. Thus they can produce generalized feature representations and can benefit a wide range of downstream tasks. For instance, CLIP-V and CLIP-T show strong performance for zero-shot classification and object detection tasks \cite{zhong2021regionclip}. Due to their complementary model architecture and massive scale pretraining, unimodal encoder models present the potential to benefit VL tasks.
However, whether and how to leverage pretrained vision/text encoder models for VL tasks is still an area of active research investigation. Previous methods extensively experimented with directly plugging in pretrained text encoders into VL frameworks \cite{wang2021simvlm,chen2020uniter,Su2020VL-BERT,gan2020large,huang2021seeing,huang2020pixel,lxmert,wang2021sgeitl}. Nevertheless, those frameworks all require redoing pretraining steps to align the representations. Another recent method \cite{shen2021clip} adapts the pretrained vision model from CLIP \cite{clip} but also has to undergo an additional stage of pretraining on millions of data for adaptation. This approach not only increases the computational complexity but also require to have sufficient additional data for adaptation. This is particularly impractical in real-world scenarios with limited data availability and may further vulnerability to domain shift in target task data.

%Thus our objective is to propose a method to effectively utilize large-scale pretrained vision, text models: (1) Into pretrained VL models without further pretraining; (2) To improve downstream VL especially highly-semantic VL tasks; (3) To robustly maintain effective even with limited or domain-shifted data. In this work, we seek to achieve the objectives and thus answer the questions in abstract through the following two contributions:

Current research is continuously pushing the performance of pretrained unimodal encoder models via increasing pretraining data size. Therefore, a natural research question rises. If given the best-pretrained vision and text encoders, what is the efficient solution to integrate them into a pretrained VL model without redoing pretraining steps and impacting inference complexity?

In this work, we propose a flexible approach to leverage pretrained vision and text models for VL tasks relying only on the task finetuning step. Specifically, we propose our main approach referred to as Multimodal Adaptive Distillation (MAD), which adaptively distills knowledge from pretrained unimodal models to VL task-specific cross-modal architectures per data instance, changing both distillation weights as well as distillation targets, depending on the behavior of the unimodal encoders and VL task-specific cross-modal architecture on the data point. Our framework, first time, proposes multimodal modularized distillation allowing teacher unimodal encoders to come from different frameworks. Our method does not require computationally expensive pretraining steps and maintains the inference complexity of the task model. For evaluation, we propose a new protocol involving multiple VL tasks (VCR, SNLI-VE, and VQA) where evaluation is conducted under a variety of data availability constraints, including zero-shot, low-shot, and fully-supervised settings. For VCR, we additionally use established ways of perturbing the characteristics of the evaluation set to measure model performance under domain-shift. We compare ours with a broad spectrum of baselines fusing pretrained vision and text models and demonstrate superior performance, achieving state-of-art on VCR Q2A for single models pretrained with image-text data, and competitive results on SNLI-VE and VQA.

\section{Related Work}
% \vspace{-1mm}
\textbf{Vision and Language Pretraining: }  Since the success of text pretraining with attention-based models, such as BERT \cite{devlin2018bert}, GPT \cite{radford2019language}, etc., those pretrained text encoder models have been broadly employed in different tasks.  VL models, such as LXMERT \cite{lxmert}, VL-BERT \cite{Su2020VL-BERT}, UNITER \cite{chen2020uniter}, VILLA \cite{gan2020large}, and others \cite{huang2021seeing,huang2020pixel,zhou2020unified,kim2021vilt,yu2020ernie,li2021align} followed to integrate the pretrained text models into their frameworks. Thus they have to conduct additional VL pretraining (VLP) with image-caption data. All of these pretraining steps utilize additional datasets on the order of 10M samples for various pretraining objectives, including Masked Language Modeling (MLM), Image-Text Matching (ITM), \textit{etc.} Similarly large-scale pretrained vision models can be obtained via vision pretraining frameworks such as ViT \cite{dosovitskiy2020image} on image dataset including JFT-300M \cite{sun2017revisiting}. Trying to utilize large-scale pretrained vision models, a recent work, CLIP-ViL experiments to \cite{shen2021clip} adapt the vision encoder from CLIP. Similarly, this also unavoidably results in redoing all the VLP steps. Both pretrained vision and text models can also be obtained via contrastive pretraining frameworks which reserve modality-independence \cite{clip,jia2021scaling,basic}. These contrastive learning frameworks can also be regarded as a shallow method to fuse pretrained vision and text encoders for VL tasks. They heavily rely on cosine distance between output features of text and vision encoder models. However, without incorporating pretrained vision, text models into a pretrained VL model but a shallow cosine measure, these frameworks have not proved to produce a high performance on highly semantic VL tasks like VCR, \textit{etc.}

Currently, there is a lack of a more general framework for fusing large-scale pretrained vision and text models into a pretrained VL structure for improving downstream VL tasks while controlling computational complexity. Existing fusion methods, such as shallow contrastive frameworks, that avoid impacting computation complexity mostly focus on traditional VL tasks such as object classification, image captioning, \textit{etc.} To our best knowledge, no prior VLP works from this tier have studied the impact on generalization capability by assessing performance on highly-semantic tasks, such as VCR, especially under both low-shot and domain shifted scenarios, which are more reflective of the challenges encountered in practice.

% CLIP \cite{clip} with 400 million, both ALIGN \cite{jia2021scaling} and SimVLM \cite{wang2021simvlm} with 10 billion and BASIC \cite{basic} with 100 billion. For this tier, with larger pretraining, their outputed feature representations are more generalized and can benefit a wide range of downstream tasks. For instance, CLIP shows strong performance for zero-shot classification tasks; SimVLM focuses on VL tasks such as VQA, SNLI-VE, and captioning. Among them, SEs from frameworks like CLIP, ALIGN and BASIC can be conveniently adapted to other tasks or model structures.

% On the other hand, it is indeed difficult to transfer pretrained knowledge from frameworks like SimVLM which lacks of SEs and constituted mainly by nested fusion layers. It is obviously impossible to believe that large-scale pretrained frameworks without SEs like SimVLM are the optimized structures for all different kinds of downstream tasks. Thus, when training certain domain-favored or task-specific models, pretrained models like SimVLM would have very low utility.

{\em In this work}, we address several of these mentioned gaps. First, we propose an efficient distillation approach to leverage pretrained vision and text encoder models into a pretrained VL structure in a way that doesn't require additional pretraining, and retains inference complexity. Second, we study the impact of utilizing large-scale pretrained vision, text encoders like CLIP-V, CLIP-T, Roberta, ViT, \textit{etc.} on VL tasks including VCR, SNLI-VE, and VQA under true zero-shot, low-shot, and domain shifted scenarios.

\textbf{Generalization of Question Answering: }Visual question answering \cite{zellers2019vcr,VQA} is among the most challenging VL tasks, due to large variation between samples and distribution discrepancy between training and testing datasets. This results in difficulty for models to perform well under zero-shot, low-shot, and domain-shifted settings. Despite high accuracy recent models achieve \cite{su2019vl,gan2020large,li2020oscar,zhang2021vinvl,zellersluhessel2021merlot}, other works have begun to uncover the tendency of models to leverage spurious shortcut signals, or memorization of mapping distributions \cite{sen2020models,jiang2019avoiding,dancette2021beyond,kovaleva2019revealing}. While prior works have begun to explore question answering tasks under zero-shot \cite{teney2016zero,noh2019transfer}, low-shot \cite{chada2021fewshotqa,brown2020language} and domain shifted \cite{dancette2021beyond,jiang2019avoiding,goyal2017making,zhang2016yin,agrawal2018don,shah2019cycle,ramakrishnan2018overcoming} settings, they are mostly limited within text-only question answering and low-level visual question answering tasks (\textit{i.e.} VQA \cite{VQA}). To the best of our knowledge, none of the prior works have explored true zero-shot and low-shot settings in highly complex visual question answering datasets such as VCR or SNLI-VE, and only one prior work \cite{debias} contributed to VCR with domain shift.

{\em In this work}, we conduct a thorough evaluation with many existing top-performing models on zero-shot, low-shot, and domain-shifted settings.

\textbf{Knowledge Distillation: }
In conventional knowledge distillation \cite{liu2020adaptive,yang2020knowledge,wang2021knowledge,mun2018learning}, a larger model serves as the teacher to a smaller student model. The goal is usually to obtain a computationally lighter and more efficient framework but still maintain similar or even higher accuracy \cite{googleod,cho2021dealing,sanh2019distilbert,jiao2019tinybert,kim2016sequence,sun2019patient}. However, recently many frameworks with small model complexity may have large pretraining data which results in more generalized feature representations. Crucial values still exist for distillation from those large-scale pretrained models to downstream domain-specific models with potentially larger model complexity. It is shallow to only compare the inference complexity between them ignoring the large pretraining complexity.

Also, tremendous progress was made in knowledge distillation with unimodal data. For instance, in vision, \cite{tian2020contrastive,fang2021seed} propose distillation for visual representation learning. In addition, remarkable advances have been made in knowledge distillation for language model compression \cite{sanh2019distilbert,jiao2019tinybert,sun2020mobilebert}. A wide range of works have explored to supervise the student model by mimicking different components of the teacher, \textit{e.g.} the distribution of self-attention, intermediate representations of transformer blocks, last layer features, \textit{etc.} to increase performances \cite{xu2020bert,sun2020contrastive,chen2021wasserstein}. On the other hand, only one prior explored knowledge distillation in VL transformers \cite{fang2021compressing} while it mainly follows the conventional methods with unimodal distillation. The distillation is not differentiated between vision and language.

Further, in most of these scenarios, the feature components leveraged for the distillation loss is fixed, and the weight of the distillation loss is also fixed. Prior works related to attention structures only focus on fixed sequence-to-sequence distillation \cite{sun2019patient} and only one prior work first introduced the idea of dynamically adjusting the distillation loss weight on a per-instance level \cite{furlanello2018born}.

%However, this concern needs to be addressed urgently since VL pretraining with smaller models are becoming popular.

{\em In this work}, our framework, for the first time, proposes multimodal distillation which modulates the distillation between modalities enabling flexible switch of different teacher models for each modality. Our approach is also model-agnostic ignoring the inference complexity comparison between the teacher and the student. The experiments prove that it is effective in all of these scenarios. We also leverage the idea of dynamic distillation weights, but we even go one step further: we dynamically adjust not only the {\em distillation weight} for each sample instance but {\em which token features are distilled} as well within every sequence.

% \vspace{-3mm}
\begin{figure*}[htpb]
\resizebox{12.2cm}{!}{%
\centering
\includegraphics[width=1\linewidth]{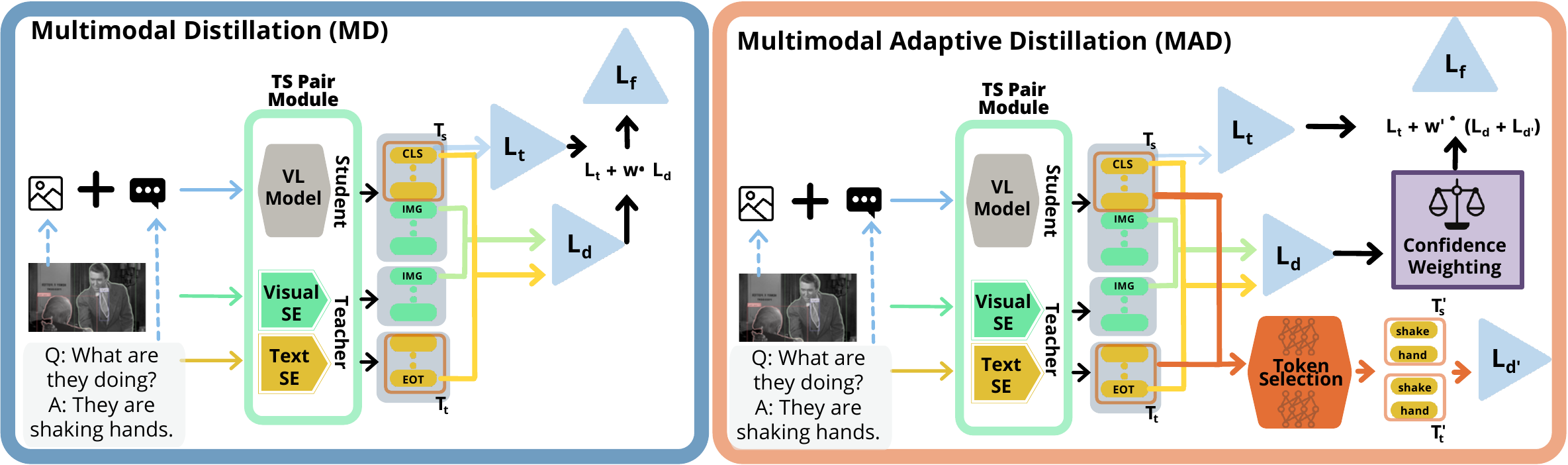}
}
\caption{Structure diagram of Multimodal Distillation (MD) and Multimodal Adaptive Distillation (MAD). MAD is further improved on top of MD and MD is equivalent to MAD in terms of structure when both Confidence Weighting and Token Selection are removed.}
% \vspace{-3mm}
\label{fig:diagram2}
\end{figure*}

% For a given task with an established base model as the student, a token selection method chooses tokens to distill from  the teacher with SEs to the student. Then confidence weighting determines the weight for this distillation instance, based on confidence of the teacher and the student. We perform thorough ablation to understand contributions of each component, and compare to naive distillation, which removes both confidence weighting and token selection.

% \begin{figure}[htpb]
% \centering
% \includegraphics[width=1\linewidth]{figures/phase.pdf}

% \caption{Comparison of training Strategies across different data domains. Our method does not impact on the training and inference complexity and only require the downstream target in-domain data.}
% % \vspace{-5mm}
% \label{fig:diagram3}
% \end{figure}
% \vspace{-6mm}

\section{Methods}
% \vspace{-2mm}
In this work, we explore many methods to utilize large-scale pretrained unimodal encoder models to help downstream VL tasks, including direct finetuning, adding adapters on unimodal models, and several proposed forms of distillation. Among these, we find that our proposed knowledge distillation is the most effective, achieving the best performance without having to redo costly pretraining, and without impacting inference complexity. We introduce these methods in the following order: First, a naive approach, Multimodal Distillation (MD); Then, the improved Multimodal Adaptive Distillation (MAD).

\subsection{Multimodal Distillation}
% \vspace{-1mm}
Following \cite{sanh2019distilbert,jiao2019tinybert,kim2016sequence,sun2019patient}, we can utilize token features form both vision and language to represent the corresponding modality sequence information. This allows our method to conduct sequence-to-sequence distillation.

Both the teacher and student model's visual $img$\footnote{visual $img$ token refer to the first token in visual sequence} tokens, in addition to the teacher's text $eos$ token and the student's text $cls$ token, are compared via L1 measure. The final loss is the weighted distillation loss $L_{d}$ summed with the original task loss $L_{t}$ for any specific downstream task. Formally,
% \vspace{-2mm}
\begin{equation}
\begin{aligned}
L_{final-MD} &=L_{t}+w \cdot\left(L_{d}\right)
\end{aligned}
\label{eq:vl}
\end{equation}
% \vspace{-1mm}

\noindent where the distillation loss $L_{d}$ is a  sum of the distillation losses between the two modalities:
% \vspace{-2mm}
\begin{equation}
\begin{aligned}
L_{d} &=L_{d,v}+L_{d,t}
\end{aligned}
\end{equation}
% \vspace{-3mm}

\noindent where $v$ and $t$ refer to vision and text branches, respectively.
% \vspace{-3mm}
\begin{equation}
\begin{aligned}
L_{d,v} &=\left\|f_{t, v,img}\left(i_{j}\right)-f_{s,v,img}\left(\partial\left(i_{j}\right)\right)\right\|_{1} = \left\| V_{t} - V_{s}\right\|_{1}
\end{aligned}
\end{equation}

\noindent where $f_{t,v,img}$ and $f_{s,v,img}$ refer to the feature extraction regarding $img$ token of the teacher vision encoder, and student model. $ V_{t}$ and $ V_{s}$ represent the extracted visual token features from the teacher and student models. $\partial$ represents the backbone detection network. $i_{j}$ refers to the image input for instance $j$. For the text modality, we have
% \vspace{-3mm}
\begin{equation}
\begin{aligned}
L_{d,t} &=\left\|f_{t,t,eos}\left(t_{j}\right)-f_{s,t,cls}\left(t_{j}\right)\right\|_{1} =\left\| T_{t} - T_{s}\right\|_{1}
\end{aligned}
\end{equation}

\noindent where $f_{t,t,eos}$ refers to the feature extraction of $eos$ token from the teacher text model, and $f_{s,t,cls}$ refers to the feature extraction of $cls$ token from the student model. $ T_{t}$ and $ T_{s}$ represent the extracted text token features from the teacher and student models. $t_{j}$ refers to the text input for instance $j$.
% \vspace{-2mm}
\subsection{Multimodal Adaptive Distillation (MAD) }
% \vspace{-1mm}
Improved on top of MD, MAD consists of the following 3 components: Token Selection using an unsupervised language prior, Confidence Weighted Distillation, and Adaptive Finetuning with Distilled Knowledge.

%and Adaptive Finetuning (AF) with Contrastive Knowledge.

{\bf \noindent Token Selection:} Typically, distillation methods \cite{sanh2019distilbert,jiao2019tinybert,kim2016sequence,sun2019patient} distill knowledge from the teacher transformer to the student transformer in a sequence-to-sequence fashion by using fixed token components of the networks. However, the most semantically relevant tokens can change {\em per instance}. Without a dynamic process to determine which components of a network should be distilled, the student model may have a higher risk of learning spurious signals from trivial tokens. Addressing this, we present a hybrid method of performing token selection for distillation.

Given a text sequence $t_{j}=\left\{w_{0} \ldots w_{z}\right\}$, where $z$ is the length of the sequence, we apply a pretrained Token Selection Module (TSM) $f_{tsm}\left(i_{j},t_{j}\right)$ to discriminate the semantically meaningful tokens. TSM generates a score, $s_{l}$ for each token $w_{l}$, obtaining a distribution $S_{j}=\left\{s_{0} \ldots s_{z}\right\} = f_{tsm}\left(i_{j}, \left\{w_{0} \ldots w_{z}\right\}\right)$. In our implementation of this selective module, two sets of weights are computed via two different approaches, and summed together:

\begin{equation}
\begin{gathered}
S_{j}=\frac{S_{vr}}{|S_{vr}|_{1}} + \frac{S_{si}}{|S_{si}|_{1}} \\
\end{gathered}
\end{equation}

\noindent where $S_{vr} = \left\{s_{vr0} \ldots s_{vrz}\right\} = \left\{cos\left<f_{t, v}\left(i_{j}\right), f_{s,t}\left(w_{l}\right)\right>\right\}$ represents a scoring between the visual representation and text token representation. In this manner, $S_{vr}$ is the score measuring the visual relevance of each token $w_{l}, l \in[0, \ldots, z]$. Then, $S_{si} = f_{ke}\left(t_{j}\right)$, where $S_{si}$ represents
the semantic and syntactic importance of each token related to the context, $t_{j}$, using a keyword extractor. In practice, we apply a pre-trained keyword extractor\cite{yake}, $f_{ke}$ with $n$-grams:

% \begin{equation}
% t_{j}^{\prime}=f_{argmax}\left(S_{j}, t_{j}, m \right)
% \end{equation}

% \noindent $f_{argmax}$ is a function that would
\noindent We then rank text tokens based on $S_{j}$ and select $m$ tokens with the highest scores, $t_{j}^{\prime}$, where $\left|t_{j}{ }^{\prime}\right| = m$. The corresponding features of both teacher and student model would be compared with an L1 measure to calculate their difference:

% \vspace{-2mm}
\begin{equation}
L_{d t}{ }^{\prime}=\left\| f_{t, t}\left( t_{j}{ }^{\prime}\right)-  f_{s, t}\left(t_{j}{ }^{\prime}\right)\right\|_{1}
\end{equation}

\noindent finally, the outputted loss, $L_{d t}{ }^{\prime}$ would be added to the final loss, $L_{\text {final-MAD }}$ by the proportionally updated distillation weight, $w^{\prime}$:
% \vspace{-1mm}
\begin{equation}
L_{\text {final-MAD}}=L_{t}+w^{\prime} \cdot\left(L_{d,v}+L_{d,t}+L_{d t}^{\prime}\right)=L_{t}+w^{\prime} \cdot\left(L_{d}+L_{d t}^{\prime}\right)
\label{eq:ts}
\end{equation}

{\bf \noindent Confidence Weighting (CW):} CLIP \cite{clip} benefits from a large amount of paired language-image training data. While this broad prior knowledge is likely helpful for VL tasks, the degree to which this knowledge is either helpful or potentially hurtful likely changes on an {\em instance level}. Given this, we design an approach to toggle the distillation objective depending on the relative confidence of the CLIP teacher and the specific student architecture. To do this, we define the ratio $r$ between maximum confidence scores of the CLIP teacher and base student model:

\begin{equation}
r= \frac{f_{argmax} \left(\sigma\left(L_{j}^{c}\right)\right)}{f_{argmax}\left(\sigma\left(L_{j}^{b}\right)\right)}
\end{equation}
% \vspace{-2mm}
\noindent where $L_{j}^{c}$ represents the logit vector from CLIP and $L_{j}^{b}$ from the base model. Finally, the new adaptive weight $w_{r}$ is defined as:
% \vspace{-1mm}
\begin{equation}
w_{r}= \begin{cases}\text {0, } & \text {if r } \leq 1 \\ w^{\prime}\text{, } & \text {if r }>1\end{cases}
\end{equation}

\noindent where $w_{r}$ replaces the distillation weights in Eq. \ref{eq:vl} and \ref{eq:ts}. When the ratio is above 1, CLIP is confident with its prediction, thus the distillation weight $w^{\prime}$ would be applied accordingly. Otherwise, the distillation value would be set to 0 to prevent from CLIP's interference.

{\bf \noindent Adaptive Finetuning (AF) : } After large-scale pretraining, previous work \cite{chen2020uniter} chooses to first train V+L models with a set of auxiliary tasks to better integrate the modalities, including Masked Language Modeling (MLM), Image-Text Matching (ITM), \textit{etc.} on downstream datasets. Then the training would finally be followed by the direct finetuning with the downstream target task only. The corresponding loss of this set of auxiliary tasks can be denoted as  $L_{\text {adapt}}$ (For details of $L_{\text {adapt}}$, please refer to \cite{chen2020uniter}). Inspired by this, we further propose a two-stage finetuning strategy. With our strategy, we first finetune the base model with $L_{\text { AF }}$ on the full downstream data then we conduct the last-step finetuning with $L_{\text {final-MAD}}$. Thus during the Adaptive Finetuning, multimodal distilled knowledge can help align the vision and language modality features beforehand.

% \vspace{-1mm}
% \begin{equation}
% L_{A F }=L_{\text {adapt}}+w \cdot\left(L_{d,v}+L_{d,t}\right)=L_{\text {adapt}}+w \cdot L_{d}
% \end{equation}
\begin{equation}
\text{Full Training Steps}= \begin{cases}\text{1st step, } & \text{finetune with }L_{\text{AF}} \\ \text{2nd step, } & \text{finetune with }L_{\text {final-MAD}} \end{cases}
\end{equation}
% \vspace{-3mm}
\subsection{Teacher Models}
We utilize the popular pretrained visual and text encoders like CLIP-V, CLIP-T, ViT and RoBERTa. Our distillation framework is a generalized modularized multimodal distillation such that the teacher model of each modality can be different. This leads to forming two essential combinations in our experiments: pretrained visual and text encoders come from the same pretraining framework and \textit{vice versa}.

% \vspace{-5mm}
\subsection{Student Models}
% \vspace{-2mm}
 Several recent top-performing pretrained VL models for highly-semantic VL tasks are selected as students, including UNITER ~\cite{chen2020uniter}, VL-BERT ~\cite{Su2020VL-BERT}, and VILLA ~\cite{gan2020large}. All of these models represent variations of multi-modal architectures, using different portfolios of objective functions and image-caption datasets for pretraining.

%  \vspace{-4mm}
 \section{Datasets and Evaluations}
In order to evaluate our methods and demonstrate their supremacy. We propose this evaluation protocol under a variety of data availability constraints, including zero-shot, low-shot, domain-shift, and fully-supervised settings. We evaluate our methods on 3 commonly used highly-semantic VL benchmarks: VCR, SNLI-VE, and VQA.

%   \vspace{-2mm}
 \subsection{Visual Commonsense Reasoning (VCR)}

The VCR benchmark presents images along with a paired question, a set of candidate answers, and a set of candidate rationales ~\cite{zellers2019vcr}. The dataset includes 290k questions, in reference to 110k unique visual scenes. The questions are constituted into 7 categories based on patterns in the questions. Please see the supplementary material for a full list.

{\bf \noindent Zero-Shot:}
No training data is used. The pretrained model is directly employed to produce a matching between the image-question pair and a candidate answer. Answers are selected based on which produce the best matches, according to the model's matching measure.

{\bf \noindent Low-Shot:}
In the low-shot setting, we have 2 training set partitions of varying sizes. Since VCR has \textbf{7 types} of questions thus (1) we select 100 examples \textbf{per question category}, totalling 700 pairs, or 0.3\% of the entire dataset, and  (2) 1,000 examples per category, totalling 7,000 pairs, or 3\%. Each experiment is run more than 4 times.

% {\bf \noindent Semi-Supervised:}
% In semi-supervised, we follow the same setting of Low-Shot for direct finetuning; however, the model has access to the remaining training data without groundtruth labels for distillation.

% {\bf \noindent Fully Supervised:}
% All question-answer pairs are used.

{\bf \noindent Standard Evaluation:} In this evaluation setting we follow the standard protocol in the benchmark.

{\bf \noindent Shortcut Mitigated (SM):}
We include a prior evaluation configuration ~\cite{debias} that focuses on mitigating shortcuts between question and answers. Shortcuts are shallow signals models can learn to recognize. Learning shortcuts may allow models to link questions to correct answers without deep understanding of the content.  We refer to this as ``Shortcut Mitigated'' (SM).

% \vspace{-4mm}

\subsection{Visual Entailment (SNLI-VE)}

% \vspace{-4mm}

The Stanford Natural Language Inference Visual Entailment (SNLI-VE) task ~\cite{xie2019visual} presents images as a premise, with paired hypothesis test. The goal is to predict whether the image entails or contradicts the hypothesis, or whether neither is the case (neutral).
% This is typically framed as a classification problem, since the predictions are fixed.

{\bf \noindent Zero-Shot:}  We first extract the VL features from the pretrained models for the given image and text premise. The features are then directly measured by cosine distance to produce the similarity between the image and the text premise. The challenge is determining what similarity values should constitute entailment, contradiction, or neither. To accomplish this, without finetuning the model, we further perform k-means of the similarities on the validation set, with $k=3$, and use the resultant clusters as anchors for each output decision. Note that this approach, while a form of transductive learning on the validation set, uses no ground truth labels, and does not change any weights of the pretrained models. We apply this procedure to evaluate both pretrained VL encoders and pretrained VL models.

{\bf \noindent Low-Shot:}
There are only 3 types of relationship between the image premise and the text hypothesis (entailment, neutral, and contradiction). Also, different from VCR, an image can pair with around 5 text premises in SNLI-VE. Therefore, we choose the image-based selection. We also have two settings: (1) We select 100 random images for each class, {\em 300} images in total paired with around {\em 1,500} text premises. (2) 1,000 randomly sampled for each class label, then {\em 30,00} samples in total with around {\em 15,000} premises. They both correspond to $0.3\%$ and $3\%$ of SNLI-VE. Each experiment is run more than 4 times.

% {\bf \noindent Fully Supervised:}
% All question-answer pairs are used for training or finetuning.

\subsection{Visual Question Answering (VQA)}

Different from VCR and VE, for every image-question pair in VQA \cite{VQA},  question-specific multiple choices are not provided. Instead, the global set of all possible answer choices for all the questions are provided (more than 3,000). The challenge is then to select the correct answer choice from this set for the given image-question pair.

% More importantly, all the answer choices for questions in VQA  consists of single words or short phrases. Thus, due to the semantic simplicity and ambiguity of the question and answer, a given question can potentially have more than one mapped answer choice.

{\bf \noindent Zero-Shot: }Comparing with VCR and VE, zero-shot on VQA is more challenging due to the large amount of answer choices (refer to supplement).

{\bf \noindent Low-Shot:}
There is not a clear categorization of VQA questions. Based on our analysis of the first n-gram words of questions, we group and finalize to 8 types in total. In VQA, an image is also paired with several questions, 5.4 on average. Thus, we also rely on image-based sampling. We have two settings of image sampling in low-shot settings: 100 random images per question types and 1,000 random image per question types. After collecting up to 2 paired questions per each selected image, we have two low-shot set: a set with around 1,600 questions and another set with 16,000 questions. They both correspond to $0.3\%$ and $3\%$ of VQA. Each experiment is run more than 4 times.

% {\bf \noindent Fully Supervised:}
% All question-answer pairs are used for training or finetuning.

\begin{table*}[]
% \small
\begin{center}
\resizebox{13cm}{!}{%
\begin{tabular}{|c|c|c|cc|cccc|cccc|lllllllllll}
\cline{1-13}
\multicolumn{1}{|l|}{}                                                             & \textbf{\begin{tabular}[c]{@{}c@{}}Base (Student)\\ Model\end{tabular}} & \textbf{Method}                      & \multicolumn{2}{c|}{\textbf{Teacher Model}}                            & \multicolumn{4}{c|}{\textbf{Standard   Evaluation}}                                                         & \multicolumn{4}{c|}{\textbf{SM   Evaluation}}                                                               &  &  &  &  &  &  &  &  &  &  &  \\ \cline{1-13}
\multirow{2}{*}{\textbf{}}                                                         & \multirow{2}{*}{}                                                       &                                      & \multicolumn{1}{c|}{VE}                      & TE                      & 0\%                  & 0.3\%                        & 3\%                           & 100\%                 & 0\%                  & 0.3\%                        & 3\%                           & 100\%                 &  &  &  &  &  &  &  &  &  &  &  \\
                                                                                   &                                                                         & \multicolumn{1}{l|}{}                & \multicolumn{1}{l|}{}                        & \multicolumn{1}{l|}{}   & \multicolumn{1}{l}{} & \multicolumn{1}{l}{100 SP/C} & \multicolumn{1}{l}{1000 SP/C} & \multicolumn{1}{l|}{} & \multicolumn{1}{l}{} & \multicolumn{1}{l}{100 SP/C} & \multicolumn{1}{l}{1000 SP/C} & \multicolumn{1}{l|}{} &  &  &  &  &  &  &  &  &  &  &  \\ \cline{1-13}
\multirow{12}{*}{\textbf{\begin{tabular}[c]{@{}c@{}}Only\\ Finetune\end{tabular}}} & -                                                                       & Direct Finetune                      & \multicolumn{1}{c|}{\multirow{2}{*}{CLIP-V}} & \multirow{2}{*}{CLIP-T} & 54.82                & 38.23                        & 38.10                         & 54.23                 & 58.30                & 34.85                        & 39.06                         & 52.23                 &  &  &  &  &  &  &  &  &  &  &  \\ \cline{2-3}
                                                                                   & -                                                                       & Adapters                             & \multicolumn{1}{c|}{}                        &                         &                      & 34.41                        & 36.34                         & 36.42                 &                      & 33.35                        & 35.02                         & 35.48                 &  &  &  &  &  &  &  &  &  &  &  \\ \cline{2-13}
                                                                                   & \multirow{3}{*}{VL-BERT}                                                & Baseline                             & \multicolumn{1}{c|}{-}                       & -                       & 23.37                & 30.85                        & 53.48                         & 75.53                 & 23.24                & 26.37                        & 49.27                         & 71.13                 &  &  &  &  &  &  &  &  &  &  &  \\ \cline{3-5}
                                                                                   &                                                                         & \textbf{MD$\star$}                   & \multicolumn{1}{c|}{\multirow{2}{*}{CLIP-V}} & \multirow{2}{*}{CLIP-T} &                      & 36.78                        & 55.91                         & 76.37                 &                      & 34.93                        & 52.06                         & 73.29                 &  &  &  &  &  &  &  &  &  &  &  \\ \cline{3-3}
                                                                                   &                                                                         & \textbf{MAD$\star$}                  & \multicolumn{1}{c|}{}                        &                         & \textbf{26.80}       & \textbf{40.43}               & \textbf{58.98}                & \textbf{77.61}        & \textbf{26.31}       & \textbf{39.27}               & \textbf{54.88}                & \textbf{74.55}        &  &  &  &  &  &  &  &  &  &  &  \\ \cline{2-13}
                                                                                   & \multirow{3}{*}{UNITER}                                                 & Baseline                             & \multicolumn{1}{c|}{-}                       & -                       & 24.78                & 31.43                        & 54.24                         & 76.67                 & 24.21                & 28.43                        & 52.72                         & 73.84                 &  &  &  &  &  &  &  &  &  &  &  \\ \cline{3-5}
                                                                                   &                                                                         & \textbf{MD$\star$}                   & \multicolumn{1}{c|}{\multirow{2}{*}{CLIP-V}} & \multirow{2}{*}{CLIP-T} &                      & 39.43                        & 57.64                         & 76.80                 &                      & 38.21                        & 54.58                         & 74.01                 &  &  &  &  &  &  &  &  &  &  &  \\ \cline{3-3}
                                                                                   &                                                                         & \textbf{MAD$\star$}                  & \multicolumn{1}{c|}{}                        &                         & \textbf{26.78}       & \textbf{42.23}               & \textbf{60.88}                & \textbf{77.05}        & \textbf{26.49}       & \textbf{41.83}               & \textbf{54.64}                & \textbf{74.24}        &  &  &  &  &  &  &  &  &  &  &  \\ \cline{2-13}
                                                                                   & \multirow{4}{*}{VILLA}                                                  & Baseline                             & \multicolumn{1}{c|}{-}                       & -                       &                      & 34.84                        & 57.01                         & 78.27                 &                      & 29.41                        & 54.15                         & 75.43                 &  &  &  &  &  &  &  &  &  &  &  \\ \cline{3-5}
                                                                                   &                                                                         & \multirow{3}{*}{\textbf{MAD$\star$}} & \multicolumn{1}{c|}{\multirow{2}{*}{CLIP-V}} & CLIP-T                  &                      & 42.95                        & 60.93                         & 78.83                 &                      & 41.97                        & 55.20                         & 76.01                 &  &  &  &  &  &  &  &  &  &  &  \\ \cline{5-5}
                                                                                   &                                                                         &                                      & \multicolumn{1}{c|}{}                        & RoBERTa                 &                      & \textbf{43.11}               & \textbf{61.49}                & \textbf{78.91}        & \textbf{}            & \textbf{41.98}               & \textbf{56.85}                & \textbf{76.32}        &  &  &  &  &  &  &  &  &  &  &  \\ \cline{4-5}
                                                                                   &                                                                         &                                      & \multicolumn{1}{c|}{ViT}                     & CLIP-T                  &                      & 40.73                        & 58.38                         & 78.59                 &                      & 39.44                        & 54.29                         & 75.35                 &  &  &  &  &  &  &  &  &  &  &  \\ \cline{1-13}
\textbf{\begin{tabular}[c]{@{}c@{}}Re-\\ Pretrain\end{tabular}}                    & CLIP-ViL$_{p}$                                                          & \textbf{}                            & \multicolumn{1}{c|}{CLIP-V}                  &                         &                      & 34.63                        & 53.54                         & 68.36                 &                      & 33.41                        & 52.44                         & 66.83                 &  &  &  &  &  &  &  &  &  &  &  \\ \cline{1-13}
\end{tabular}
}
\end{center}
\caption{VCR dataset results, including approaches for direct finetuning of adapters, and distillation to various student architectures. $\star$ represents our methods. Although we avoid extra pre-training, we include a recent top-performing method utilizing CLIP vision encoder with additional pre-training here for comparison (CLIP-ViL$_p$). Data sampling shown as number of image-question-answer triplets (0-shot to Full). Few-shot results are averaged over 4 runs. Refer to supplement for standard deviations, details of 0-shot experiments, and sizes of student models. Baseline refers to the original methods without distillation. Their results are based on our re-implementation of the students models with additional hyper-parameter tuning. SM = Shortcut Mitigated. }
%  \vspace{-6mm}
\label{tab:vcr}
%  \vspace{-2mm}
\end{table*}

%  \vspace{-3mm}

\section{Results}

\subsection{Visual Commonsense Reasoning (VCR)}

Results on the VCR dataset for Q$\rightarrow$A, across several data availability and domain, shifted constraints, are shown in Tab. \ref{tab:vcr}  (for additional metrics of Q2A and Q2AR, please see supplement). These results yield 5 key observations: 1) From the 1st row, pretrained unimodal encoder models are capable of strong zero-shot results in VCR ``out-of-the-box'': When TE and VE are both CLIP's vision and text encoders, the accuracy is over 58\% on short-cut mitigated evaluation (a more difficult domain-shifted task), which is similar to some supervised approaches. 2) Based on the 2nd row, without enough downstream data to finetune, pretrained unimodal encoders may not adapt to the downstream domain well. In fact, new data may disturb the pretrained knowledge and result in even more poor performance than zero-shot.3) Our proposed MAD approach yields significant performance gains under low-shot data regimes across a range of student architectures: up to 52\% for VL-BERT, and 47.7\% for UNITER. 4) Our proposed MAD approach even benefits fully supervised tasks by 2.1\% for VL-BERT and 3.8\% for UNITER. 5) Our approach yields even higher gains under low-shot domain-shifted scenarios of shortcut mitigation (SM): up to 71.3\% for VL-BERT and 61\% for UNITER. 6) With only finetuning and not increasing any computation complexity for the student model, MAD outperforms a prior proposed approach to leverage CLIP's vision encoder and redo the pretraining steps: CLIP-Vil \cite{shengshen}. 7) The experiments prove that our modularized multimodal distillation method is generalizable and flexible. It can maintain effectiveness regardless of whether the pretrained vision and text encoders come from the same pretraining framework or separate ones.

MAD with VILLA delivers high performance on the public leaderboard (Q2A: \textbf{79.6\%} QA2R: \textbf{82.9\%} Q2AR: \textbf{66.2\%}), achieving a new state-of-art of Q2A performance compared to other single models that are pretrained with image-text data, as well as overall state-of-art comparable performance for QA2R and Q2AR.

As our approach is model agnostic, ensembles are possible. Combining multiple MAD approaches, further significant gains in performance (Q2A: \textbf{80.93\%}, QA2R: \textbf{84.01\%}) in the fully-sampled standard validation set.

\begin{table*}[]
% \tiny
\centering
\resizebox{\linewidth}{!}{%
\begin{tabular}{|c|ccccc|ccc|ccc|}
\hline
\textbf{Method}                           & \multicolumn{1}{c|}{\textbf{V.}} & \multicolumn{1}{c|}{\textbf{L.}} & \multicolumn{1}{c|}{\textbf{TS}} & \multicolumn{1}{c|}{\textbf{CW}} & \textbf{AF} & \multicolumn{3}{c|}{\textbf{Standard   Evaluation}} & \multicolumn{3}{c|}{\textbf{SM Evaluation}} \\ \hline
                                          &                                  &                                  &                                  &                                  &             & 0.3\%            & 3\%               & 100\%        & 0.3\%         & 3\%            & 100\%      \\
                                          &                                  &                                  &                                  &                                  &             & 100 SP/C         & 1000 SP/C         &              & 100 SP/C      & 1000 SP/C      &            \\ \hline
\textbf{Baseline}                         & -                                & -                                & -                                & -                                & -           & 30.85            & 53.48             & 75.53        & 26.37         & 49.27          & 71.13      \\ \cline{1-1}
\textbf{Unimodal Distillation}            & Y                                &                                  &                                  &                                  &             & 34.24            & 54.53             & 75.92        & 31.11         & 51.16          & 72.24      \\ \cline{1-1}
\textbf{Multimodal Distillation}          & Y                                & Y                                &                                  &                                  &             & 36.78            & 55.91             & 76.37        & 34.93         & 52.06          & 73.29      \\ \cline{1-1}
\textbf{}                                 & Y                                & Y                                & Y                                &                                  &             & 37.21            & 57.64             & 76.83        & 36.14         & 53.33          & 73.78      \\ \cline{1-1}
\textbf{}                                 & Y                                & Y                                & Y                                & Y                                &             & 38.28            & 58.68             & 77.15        & 37.82         & 54.12          & 74.24      \\ \cline{1-1}
\textbf{Multimodal Adaptive Distillation} & Y                                & Y                                & Y                                & Y                                & Y           & 40.43            & 58.98             & 77.61        & 39.27         & 54.88          & 74.55      \\ \hline
\end{tabular}
}
\vspace{3mm}
\caption{VCR distillation ablation experiments using VL-BERT student model. Both vision and text teacher encoder models come from CLIP of ViT-B16. Results of low-shot experiments are averaged over more than 4 times (see supplement for standard deviations). L. represents text distillation and V. represents Vision distillation. SM = Shortcut Mitigated. }
\label{tab:vcrablation}
% \vspace{-7mm}
\end{table*}

% \vspace{-3mm}
As shown in Table \ref{tab:vcrablation}, we performed an ablation study of individual components of our MD and MAD frameworks, evaluated with VL-BERT as the student model. These results demonstrate 3 key observations: 1) Language distillation, not vision (as studied in previous works), contributes the most performance improvement. However, distilling both vision and language perform better than either one alone. 2) Multimodal distillation with an adaptive mechanism like token selection, and confidence weighting can produce better performance than naive sequence-to-sequence multimodal distillation.

\begin{figure}[th!]
% \vspace{-4mm}
\begin{center}
\scriptsize
\resizebox{12cm}{!}{%
 \includegraphics[]{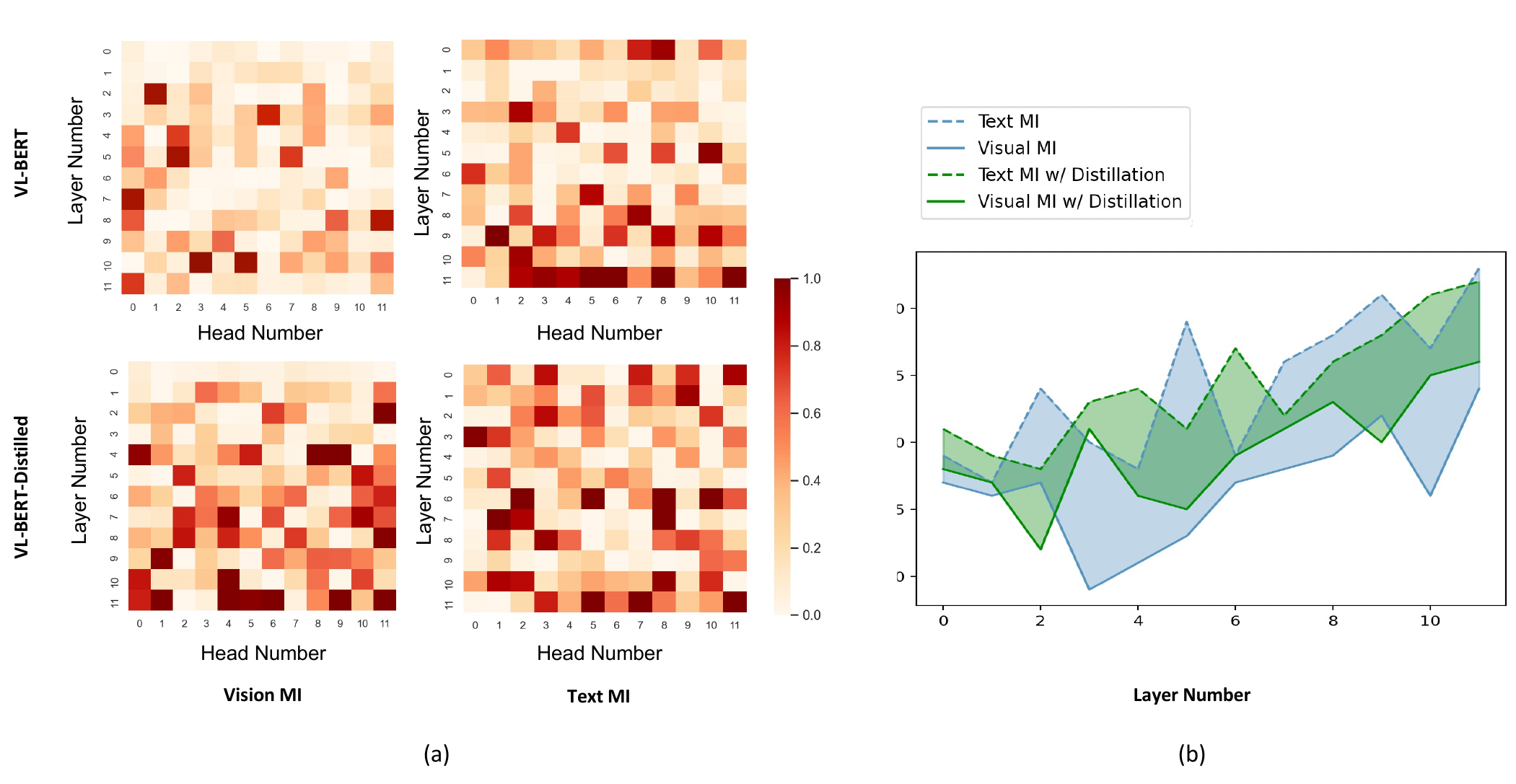}
}
\end{center}
% \vspace{-6mm}%Put here to reduce too much white space after your table
 \caption{(a) Heatmap of MI values for both visual and textual modality. Each cell represents a combination between 12 heads and 12 layers in VL-BERT Base model. The top row corresponds to the original VL-BERT Base model and the bottom row corresponds to VL-BERT Base with Multimodal Distillation. Index starts with 0. (b) Average Modality Importance (MI) values for each layer of VL-BERT with MAD (green) and baseline (blue). Shaded areas represent differences between vision and text modalities.}
\label{fig:mi_line}
% \vspace{-4mm}%Put here to reduce too much white space after your table
\end{figure}

\begin{table*}[h!]
% \vspace{-3mm}
% \tiny
\centering
\resizebox{10cm}{!}{%
\begin{tabular}{|c|c|c|cc|cccc|c|}
\hline
\multicolumn{1}{|l|}{}                                                             & \textbf{\begin{tabular}[c]{@{}c@{}}Base (Student)\\ Model\end{tabular}} & \textbf{Method}                      & \multicolumn{2}{c|}{\textbf{Teacher Model}}                            & \multicolumn{4}{c|}{\textbf{Validation}}                          & \textbf{Test}  \\ \hline
\multicolumn{1}{|l|}{}                                                             &                                                                         &                                      & \textbf{VE}                                  & \textbf{TE}             & 0-shot         & 3,000          & 30,000         & Full           & Full           \\ \hline
\multirow{12}{*}{\textbf{\begin{tabular}[c]{@{}c@{}}Only\\ Finetune\end{tabular}}} & \textbf{-}                                                              & Direct Finetune                      & \multicolumn{1}{c|}{\multirow{2}{*}{CLIP-V}} & \multirow{2}{*}{CLIP-T} & 47.23          & 50.91          & 57.02          & 66.91          &                \\ \cline{2-3}
                                                                                   & \textbf{-}                                                              & Adapter                              & \multicolumn{1}{c|}{}                        &                         &                & 38.10          & 41.65          & 41.75          &                \\ \cline{2-10}
                                                                                   & VL-BERT                                                                 & Baseline                             & \multicolumn{1}{c|}{-}                       & -                       & 33.31          & 53.28          & 62.31          & 74.66          & 74.02          \\ \cline{3-5}
                                                                                   &                                                                         & \textbf{MD$\star$}                   & \multicolumn{1}{c|}{\multirow{2}{*}{CLIP-V}} & \multirow{2}{*}{CLIP-T} &                & 56.02          & 64.92          & 75.08          & 74.93          \\ \cline{3-3}
                                                                                   & \multicolumn{1}{l|}{}                                                   & \textbf{MAD$\star$}                  & \multicolumn{1}{c|}{}                        &                         & \textbf{35.18} & \textbf{56.78} & \textbf{65.37} & \textbf{75.75} & 75.43          \\ \cline{2-10}
                                                                                   & UNITER                                                                  & Baseline                             & \multicolumn{1}{c|}{-}                       & -                       & 32.23          & 58.36          & 66.23          & 79.02          & 79.19          \\ \cline{3-5}
                                                                                   &                                                                         & \textbf{MD$\star$}                   & \multicolumn{1}{c|}{\multirow{2}{*}{CLIP-V}} & \multirow{2}{*}{CLIP-T} &                & 58.95          & 67.74          & 80.08          & 80.16          \\ \cline{3-3}
                                                                                   &                                                                         & \textbf{MAD$\star$}                  & \multicolumn{1}{c|}{}                        &                         & \textbf{33.67} & \textbf{59.42} & \textbf{68.34} & \textbf{80.14} & 80.23          \\ \cline{2-10}
                                                                                   & VILLA                                                                   & Baseline                             & \multicolumn{1}{c|}{-}                       & -                       & 34.72          & 58.47          & 67.16          & 79.64          & 79.32          \\ \cline{3-5}
                                                                                   &                                                                         & \multirow{3}{*}{\textbf{MAD$\star$}} & \multicolumn{1}{c|}{\multirow{2}{*}{CLIP-V}} & CLIP-T                  & \textbf{36.10} & \textbf{59.65} & 68.43          & \textbf{80.67} & \textbf{80.32} \\ \cline{5-5}
                                                                                   &                                                                         &                                      & \multicolumn{1}{c|}{}                        & RoBERTa                 &                & 58.48          & \textbf{68.86} & 80.64          & 80.31          \\ \cline{4-5}
                                                                                   &                                                                         &                                      & \multicolumn{1}{c|}{ViT}                     & CLIP-T                  &                & 58.08          & 66.12          & 78.37          & 78.49          \\ \hline
\textbf{\begin{tabular}[c]{@{}c@{}}Re-\\ Pretrain\end{tabular}}                    & CLIP-ViL$_{p}$                                                          &                                      & \multicolumn{1}{c|}{CLIP-V}                  &                         &                & 59.48          & 68.32          & 80.61          & 80.2           \\ \hline
\end{tabular}
}
\vspace{4mm}
\caption{SNLI-VE Results. Baselines represent the original methods without distillation. $\star$ represents our methods. Training data subsampling shown under validation results. The
low-shot experiment results are averaged over 4 runs.}
\label{tab:snlive}
% \vspace{-2mm}
\end{table*}

\begin{table*}[h!]
% \tiny

\centering
\resizebox{12cm}{!}{%
\begin{tabular}{lccccccccc}
\hline
\multicolumn{1}{|l|}{}                                                                                   & \multicolumn{1}{c|}{\textbf{\begin{tabular}[c]{@{}c@{}}Base (Student)\\ Model\end{tabular}}} & \multicolumn{1}{c|}{\textbf{Method}}                      & \multicolumn{2}{c|}{\textbf{Teacher Model}}                                                 & \multicolumn{3}{c|}{\textbf{Validation}}                                                           & \multicolumn{1}{c|}{\textbf{Test-Val}} & \multicolumn{1}{c|}{\textbf{Test-Std}} \\ \hline
\multicolumn{1}{|l|}{\multirow{2}{*}{}}                                                                  &                                                                                              & \multicolumn{1}{c|}{}                                     & \multicolumn{1}{c|}{\textbf{VE}}             & \multicolumn{1}{c|}{TE}                      & 0.3\%                        & 3\%                           & \multicolumn{1}{c|}{100\%}          & \multicolumn{2}{c|}{\multirow{2}{*}{}}                                          \\ \cline{6-8}
\multicolumn{1}{|l|}{}                                                                                   & \multicolumn{1}{l}{}                                                                         & \multicolumn{1}{l|}{}                                     & \multicolumn{1}{l|}{}                        & \multicolumn{1}{l|}{}                        & \multicolumn{1}{l}{100 SP/C} & \multicolumn{1}{l}{1000 SP/C} & \multicolumn{1}{l|}{}               & \multicolumn{2}{c|}{}                                                           \\ \hline
\multicolumn{1}{|c|}{\multirow{10}{*}{\textbf{\begin{tabular}[c]{@{}c@{}}Only\\ Finetune\end{tabular}}}} &                                                                                              & \multicolumn{1}{c|}{Adapter}                              & \multicolumn{1}{c|}{CLIP-V}                  & \multicolumn{1}{c|}{CLIP-T}                  & 16.14                        & 37.41                         & \multicolumn{1}{c|}{51.34}          & \multicolumn{1}{c|}{}                  & \multicolumn{1}{c|}{}                  \\ \cline{2-10}
\multicolumn{1}{|c|}{}                                                                                   & \multicolumn{1}{c|}{VL-BERT}                                                                 & \multicolumn{1}{c|}{Baseline}                             & \multicolumn{1}{c|}{-}                       & \multicolumn{1}{c|}{-}                       & 35.33                        & 63.29                         & \multicolumn{1}{c|}{69.05}          & \multicolumn{1}{c|}{71.79}             & \multicolumn{1}{c|}{72.22}             \\ \cline{3-5}
\multicolumn{1}{|c|}{}                                                                                   & \multicolumn{1}{c|}{}                                                                        & \multicolumn{1}{c|}{\textbf{MD$\star$}}                   & \multicolumn{1}{c|}{\multirow{2}{*}{CLIP-V}} & \multicolumn{1}{c|}{\multirow{2}{*}{CLIP-T}} & 36.83                        & 64.43                         & \multicolumn{1}{c|}{70.22}          & \multicolumn{1}{c|}{}                  & \multicolumn{1}{c|}{}                  \\ \cline{3-3}
\multicolumn{1}{|c|}{}                                                                                   & \multicolumn{1}{c|}{}                                                                        & \multicolumn{1}{c|}{\textbf{MAD$\star$}}                  & \multicolumn{1}{c|}{}                        & \multicolumn{1}{c|}{}                        & \textbf{37.12}               & \textbf{65.71}                & \multicolumn{1}{c|}{\textbf{71.42}} & \multicolumn{1}{c|}{}                  & \multicolumn{1}{c|}{}                  \\ \cline{2-10}
\multicolumn{1}{|c|}{}                                                                                   & \multicolumn{1}{c|}{UNITER}                                                                  & \multicolumn{1}{c|}{Baseline}                             & \multicolumn{1}{c|}{-}                       & \multicolumn{1}{c|}{-}                       & 36.46                        & 64.43                         & \multicolumn{1}{c|}{71.26}          & \multicolumn{1}{c|}{73.82}             & \multicolumn{1}{c|}{74.02}             \\ \cline{3-5}
\multicolumn{1}{|c|}{}                                                                                   & \multicolumn{1}{c|}{}                                                                        & \multicolumn{1}{c|}{\textbf{MAD$\star$}}                  & \multicolumn{1}{c|}{CLIP-V}                  & \multicolumn{1}{c|}{CLIP-T}                  & \textbf{39.75}               & \textbf{65.93}                & \multicolumn{1}{c|}{\textbf{71.94}} & \multicolumn{1}{c|}{}                  & \multicolumn{1}{c|}{}                  \\ \cline{2-10}
\multicolumn{1}{|c|}{}                                                                                   & \multicolumn{1}{c|}{VILLA}                                                                   & \multicolumn{1}{c|}{Baseline}                             & \multicolumn{1}{c|}{-}                       & \multicolumn{1}{c|}{-}                       & 37.18                        & 65.75                         & \multicolumn{1}{c|}{72.11}          & \multicolumn{1}{c|}{74.69}             & \multicolumn{1}{c|}{74.87}             \\ \cline{3-5}
\multicolumn{1}{|c|}{}                                                                                   & \multicolumn{1}{c|}{\textbf{}}                                                               & \multicolumn{1}{c|}{\multirow{3}{*}{\textbf{MAD$\star$}}} & \multicolumn{1}{c|}{\multirow{2}{*}{CLIP-V}} & \multicolumn{1}{c|}{CLIP-T}                  & \textbf{40.16}               & \textbf{66.93}                & \multicolumn{1}{c|}{\textbf{73.02}} & \multicolumn{1}{c|}{\textbf{75.81}}    & \multicolumn{1}{c|}{\textbf{76.04}}    \\ \cline{5-5}
\multicolumn{1}{|c|}{}                                                                                   & \multicolumn{1}{c|}{\textbf{}}                                                               & \multicolumn{1}{c|}{}                                     & \multicolumn{1}{c|}{}                        & \multicolumn{1}{c|}{RoBERTa}                 & 39.04                        & 66.23                         & \multicolumn{1}{c|}{72.43}          & \multicolumn{1}{c|}{}                  & \multicolumn{1}{c|}{}                  \\ \cline{4-5}
\multicolumn{1}{|c|}{}                                                                                   & \multicolumn{1}{c|}{\textbf{}}                                                               & \multicolumn{1}{c|}{}                                     & \multicolumn{1}{c|}{ViT}                     & \multicolumn{1}{c|}{CLIP-T}                  & 38.79                        & 66.10                         & \multicolumn{1}{c|}{72.02}          & \multicolumn{1}{c|}{}                  & \multicolumn{1}{c|}{}                  \\ \hline
\multicolumn{1}{|c|}{\textbf{\begin{tabular}[c]{@{}c@{}}Re-\\ Pretrain\end{tabular}}}                    & \multicolumn{1}{c|}{CLIP-ViL}                                                                & \multicolumn{1}{c|}{\textbf{}}                            & \multicolumn{1}{l|}{CLIP-V}                  & \multicolumn{1}{c|}{}                        & 39.01                        & 66.84                         & \multicolumn{1}{c|}{73.91}          & \multicolumn{1}{c|}{76.48}             & \multicolumn{1}{c|}{76.70}             \\ \hline
                                                                                                         & \multicolumn{1}{l}{}                                                                         & \multicolumn{1}{l}{}                                      &                                              & \multicolumn{1}{l}{}                         & \multicolumn{1}{l}{}         & \multicolumn{1}{l}{}          & \multicolumn{1}{l}{}                & \multicolumn{1}{l}{}                   & \multicolumn{1}{l}{}                   \\
                                                                                                         & \multicolumn{1}{l}{}                                                                         & \multicolumn{1}{l}{}                                      &                                              & \multicolumn{1}{l}{}                         & \multicolumn{1}{l}{}         & \multicolumn{1}{l}{}          & \multicolumn{1}{l}{}                & \multicolumn{1}{l}{}                   & \multicolumn{1}{l}{}                   \\
                                                                                                         & \multicolumn{1}{l}{}                                                                         & \multicolumn{1}{l}{}                                      &                                              & \multicolumn{1}{l}{}                         & \multicolumn{1}{l}{}         & \multicolumn{1}{l}{}          & \multicolumn{1}{l}{}                & \multicolumn{1}{l}{}                   & \multicolumn{1}{l}{}                   \\
                                                                                                         & \multicolumn{1}{l}{}                                                                         & \multicolumn{1}{l}{}                                      &                                              & \multicolumn{1}{l}{}                         & \multicolumn{1}{l}{}         & \multicolumn{1}{l}{}          & \multicolumn{1}{l}{}                & \multicolumn{1}{l}{}                   & \multicolumn{1}{l}{}                   \\
                                                                                                         & \multicolumn{1}{l}{}                                                                         & \multicolumn{1}{l}{}                                      &                                              & \multicolumn{1}{l}{}                         & \multicolumn{1}{l}{}         & \multicolumn{1}{l}{}          & \multicolumn{1}{l}{}                & \multicolumn{1}{l}{}                   & \multicolumn{1}{l}{}                   \\
                                                                                                         & \multicolumn{1}{l}{}                                                                         & \multicolumn{1}{l}{}                                      & \textbf{}                                    & \multicolumn{1}{l}{}                         & \multicolumn{1}{l}{}         & \multicolumn{1}{l}{}          & \multicolumn{1}{l}{}                & \multicolumn{1}{l}{}                   & \multicolumn{1}{l}{}                   \\
                                                                                                         & \multicolumn{1}{l}{}                                                                         & \multicolumn{1}{l}{}                                      & \textbf{}                                    & \multicolumn{1}{l}{}                         & \multicolumn{1}{l}{}         & \multicolumn{1}{l}{}          & \multicolumn{1}{l}{}                & \multicolumn{1}{l}{}                   & \multicolumn{1}{l}{}                   \\
                                                                                                         & \multicolumn{1}{l}{}                                                                         & \multicolumn{1}{l}{}                                      & \textbf{}                                    & \multicolumn{1}{l}{}                         & \multicolumn{1}{l}{}         & \multicolumn{1}{l}{}          & \multicolumn{1}{l}{}                & \multicolumn{1}{l}{}                   & \multicolumn{1}{l}{}                   \\
                                                                                                         & \multicolumn{1}{l}{}                                                                         & \multicolumn{1}{l}{}                                      & \textbf{}                                    & \multicolumn{1}{l}{}                         & \multicolumn{1}{l}{}         & \multicolumn{1}{l}{}          & \multicolumn{1}{l}{}                & \multicolumn{1}{l}{}                   & \multicolumn{1}{l}{}
\end{tabular}
}
\vspace{-20mm}
\caption{VQA Results. Baselines represent the original methods without distillation. $\star$ represents our methods. Training data subsampling shown under validation results. Test results were accomplished using full training set. The
low-shot experiment results are averaged over 4 runs.}
\label{tab:vqa}
% \vspace{-6mm}
\end{table*}
% \vspace{-8mm}

\subsection{Visual Entailment (SNLI-VE)}

% \vspace{-1mm}
Results on SNLI-VE are shown in Table \ref{tab:snlive}, revealing 3 key observations: 1) Pretrained unimodal encoders can achieve up to 40.02 \% zero-shot accuracy on val set of SNLI-VE ``out-of-the-box'', which is significantly higher than baseline methods like VL-BERT and UNITER. 2) MAD continues to provide accuracy improvements in all settings of zero-shot, low-shot, and fully-supervised conditions, across all student base models, versus both baselines with no distillation, and MD approach. 3) MAD continues to outperform concurrent work CLIP-ViL$_{p}$ under all evaluated conditions.

% With MAD, base models like VL-BERT or UNITER Base can realize improvement of 6.5 \% with only 3,000 samples and 4.9 \% with 30,000. In addition, SE-TD consistently brings improvement to all base models, demonstrating that this approach is model agnostic. In conjunction with VILLA, SE-TD can push the performance to 80.67 \% on validation and 80.31 \% on test, both of which outperform CLIP-VIL \cite{shengshen}.

% \vspace{-2mm}
\subsection{Visual Question Answering (VQA)}
% \vspace{-2mm}

Table \ref{tab:vqa} shows our results from VQA. The key observations from these experiments are as follows: 1) MAD yields performance improvement against baseline and MD under all conditions. 2) Under low-shot conditions, MAD can outperform finetuning approach CLIP-ViL$_{p}$  3) Under the full-shot setting, MAD can assist the baseline method, VILLA to reach the performance of 73.02 on validation, which is comparable to the performance from CLIP-VIL$_{p}$. Further experimentation leveraging the model-agnostic MAD to create ensembles achieves \textbf{73.93 \%} on local validation, outperforming CLIP-ViL$_{p}$.

\subsection{Analysis}

\noindent\textbf{Increased Utilization of Vision Modality: }Following \cite{cao2020behind}, we measure Modality Importance (MI) of both modalities. After distillation, refer to Fig. \ref{fig:mi_line}, we observe that the Vision MI is increased in (a) and the MI difference between Vision and Text is decreased in (b).

\begin{figure}[htpb!]
% \vspace{-4mm}
\begin{center}
\scriptsize
\scalebox{1.0}{
  \includegraphics[width=1\linewidth]{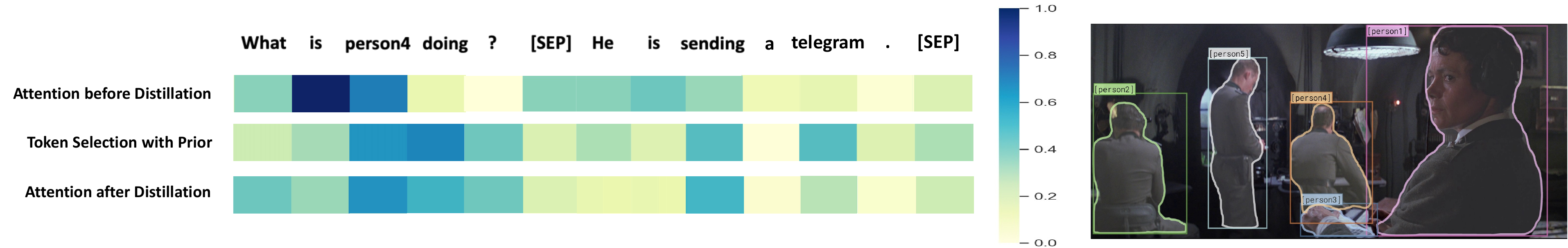}
}
\end{center}
%  \vspace{-4mm}
  \caption{Comparison of attentions from VL-BERT before and after distillation, and Token Selective Module scores.}
\label{fig:sample_example}
% \vspace{-5mm}%Put here to reduce too much white space after your table
\end{figure}

\noindent\textbf{Regulating Shortcuts: }Additionally, as V+L models are prone to learn trivial shortcuts from questions to correct answers \cite{kovaleva2019revealing,debias}, performance degrades when tested on datasets that mitigate these signals. We further seek to qualitatively better understand how our distillation approach might help improve the performance in this setting. Fig. \ref{fig:sample_example}, shows token attention values from a VL-BERT model on an instance from the VCR dataset, before and after distillation, in addition to the token selection scores of our MAD approach. One can see that trivial tokens such as ``is'' have the highest attention values prior to distillation. By contrast, the token selection forces emphasis on more meaningful terms, such as ``sending,'' ``telegram,'' \textit{etc.}

% \vspace{-5mm}

\section{Conclusions}
% \vspace{-3mm}

In this work, we explore leveraging pretrained unimodal encoders to improve existing pretrained multi-modal encoders for VL tasks. To accomplish this we present a new approach, Multi-modal Adaptive Distillation (MAD), to perform adaptive distillation from pretrianed unimodal encoders to a variety of high performing multi-modal student encoders, including VL-BERT, UNITER, and VILLA. We evaluate this approach over a new comprehensive VL task protocol involving VCR, SNLI-VE, and VQA, covering zero-shot, low-shot, fully-supervised, and domain-shifted settings. Our results demonstrate significant improvements in performance across datasets, tasks, and settings, compared with baselines.

% Compared to concurrent fine-tuning work CLIP-ViL, our approach outperforms under all low-shot conditions and most fully-supervised conditions, except for fully-supervised VQA, where the performance is comparable. Because our approach is agnostic to the base student model, ensembles of many student models can yield superior results under all conditions studied. We believe our results can be helpful for the community to know how to best leverage CLIP to potentially boost the performance of future works on VL tasks.

% \paragraph{Acknowledgement}
% Thanks to Liunian Harold Li for his help in the implementation of CLIP-ViL and feedbacks of the idea.

% \clearpage\mbox{}Page \thepage\ of the manuscript.
% \clearpage\mbox{}Page \thepage\ of the manuscript.

% This is the last page of the manuscript.
% Now we have reached the maximum size of the ECCV 2022 submission (excluding references).
% References should start immediately after the main text, but can continue on p.15 if needed.

\clearpage
\bibliographystyle{splncs04}
\bibliography{egbib}

\clearpage

\newcommand{\beginsupplement}{%
        \setcounter{table}{0}
        \renewcommand{\thetable}{S\arabic{table}}%
        \setcounter{figure}{0}
        \renewcommand{\thefigure}{S\arabic{figure}}%
     }

\renewcommand{\thesubsection}{\thesection.\alph{subsection}}
\renewcommand{\thesubsubsection}{\thesubsection.\arabic{subsection}}

% \section{Supplementary Material}
\beginsupplement

\begin{center}
\textbf{\large Supplemental Materials}
\end{center}
%  \part{}

\section{Overview}

 Section \ref{sec:vcr} covers additional details regarding our evaluations on the VCR dataset, including our public leaderboard results (\ref{sec:vcr-publicresults}), additional metrics for internal validation results (\ref{sec:vcr-valresults}, \ref{sec:vcr-sm-valresults}), information about question sub-types (\ref{sec:vcr-subtypes}), and further analysis of model behavior before and after distillation (\ref{sec:vcr-analysis}).

Section \ref{sec:trainingdetails} provides additional details regarding training code configurations (\ref{sec:trainingdetails-baselines}) and parameters in all of our experiments (\ref{sec:trainingdetails-implementation}).

Section \ref{sec:baselinemethod} covers additional details regarding our baselines for Multimodal Distillation (MD), Multimodal Adaptive Distillation (MAD) (\ref{sec:baselinemethod-nkd}) and adapters over CLIP (\ref{sec:baselinemethod-adapters}).

 \begin{table*}[h!]
\centering
\resizebox{8cm}{!}{%
\begin{tabular}{|c|c|l|ccc|}
\hline
\textbf{}                                                                          & \textbf{\begin{tabular}[c]{@{}c@{}}Base (Student)\\ Model\end{tabular}} & \multicolumn{1}{c|}{\textbf{Method}}             & \multicolumn{3}{c|}{\textbf{Standard Evaluation}}                                      \\ \hline
\textbf{}                                                                          & \textbf{}                                                               &                                                  & \multicolumn{1}{c|}{\textbf{Q2A}} & \multicolumn{1}{c|}{\textbf{QA2R}} & \textbf{Q2AR} \\ \hline
\textbf{}                                                                          & \textbf{-}                                                              & Prompt w/ zero-shot. IA(R)                       & \multicolumn{1}{c|}{54.82}        & \multicolumn{1}{c|}{48.58}         & 26.63         \\ \hline
\multirow{13}{*}{\textbf{\begin{tabular}[c]{@{}c@{}}Only\\ Finetune\end{tabular}}} & \textbf{}                                                               & Direct Finetune                                  & \multicolumn{1}{c|}{}             & \multicolumn{1}{c|}{}              &               \\ \cline{2-6}
                                                                                   & \textbf{}                                                               & Adapters                                         & \multicolumn{1}{c|}{}             & \multicolumn{1}{c|}{}              &               \\ \cline{2-6}
                                                                                   & \textbf{VL-BERT}                                                        & Baseline                                         & \multicolumn{1}{c|}{76.02}        & \multicolumn{1}{c|}{78.31}         & 59.53         \\ \cline{2-6}
                                                                                   & \textbf{}                                                               & MD $\star$                                       & \multicolumn{1}{c|}{76.74}        & \multicolumn{1}{c|}{78.64}         & 60.35         \\ \cline{2-6}
                                                                                   & \textbf{}                                                               & MAD $\star$                                             & \multicolumn{1}{c|}{77.24}        & \multicolumn{1}{c|}{79.02}         & 61.04         \\ \cline{2-6}
                                                                                   & \textbf{UNITER}                                                         & B Baseline                                       & \multicolumn{1}{c|}{74.23}        & \multicolumn{1}{c|}{76.99}         & 57.15         \\ \cline{2-6}
                                                                                   & \textbf{}                                                               & B MD $\star$                                             & \multicolumn{1}{c|}{75.21}        & \multicolumn{1}{c|}{77.59}         & 58.36         \\ \cline{2-6}
                                                                                   & \textbf{}                                                               & B MAD $\star$                                    & \multicolumn{1}{c|}{76.35}        & \multicolumn{1}{c|}{77.84}         & 59.43         \\ \cline{2-6}
                                                                                   & \textbf{}                                                               & L Baseline                                       & \multicolumn{1}{c|}{76.67}        & \multicolumn{1}{c|}{79.98}         & 61.32         \\ \cline{2-6}
                                                                                   & \textbf{}                                                               & L MAD $\star$                                    & \multicolumn{1}{c|}{77.05}        & \multicolumn{1}{c|}{80.57}         & 62.08         \\ \cline{2-6}
                                                                                   & \textbf{VILLA}                                                          & Baseline                                         & \multicolumn{1}{c|}{78.27}        & \multicolumn{1}{c|}{82.33}         & 64.44         \\ \cline{2-6}
                                                                                   & \textbf{}                                                               & MAD $\star$                                      & \multicolumn{1}{c|}{78.86}        & \multicolumn{1}{c|}{82.57}         & 65.11         \\ \cline{2-6}
                                                                                   & \textbf{Ensemble}                                                       & MAD $\star$                                      & \multicolumn{1}{c|}{80.48}        & \multicolumn{1}{c|}{82.68}         & 66.54         \\ \hline
\textbf{\begin{tabular}[c]{@{}c@{}}Re-\\ Pretrain\end{tabular}}                    & \textbf{}                                                               & CLIP-ViL$_{p}$ \cite{shengshen} & \multicolumn{1}{c|}{68.36}        & \multicolumn{1}{c|}{71.40}         & 48.81         \\ \hline
\end{tabular}%
}
\caption{Complete VCR Evaluation Results. In the first row, we utilize pretrained unimodal encoders of CLIP-T and CLIP-V to conduct zero-shot evaluation. IA represents that we disregard question infor and only we measure the cosine distance between image feature and answer feature when solving Q2A task. IR represents similar procedure between image and rationale features. $\star$ represents evaluation with our methods.}
\label{vcr_full}
 \end{table*}

\begin{table*}[h!]
\centering
\resizebox{\textwidth}{!}{%
\begin{tabular}{|ccc|c|cccccccc|}
\hline
\multicolumn{1}{|c|}{\textbf{Method}}                     & \multicolumn{2}{c|}{\textbf{Teacher Model}}                            & \textbf{Variation}    & \multicolumn{4}{c|}{\textbf{Standard   Evaluation}}                                                               & \multicolumn{4}{c|}{\textbf{SM   Evaluation}}                                                \\ \hline
\multicolumn{1}{|c|}{}                                    & \multicolumn{1}{c|}{VE}                      & TE                      & \textbf{Prompt}       & \multicolumn{1}{c|}{0\%}   & \multicolumn{1}{c|}{0.3\%} & \multicolumn{1}{c|}{3\%}   & \multicolumn{1}{c|}{100\%} & \multicolumn{1}{c|}{0\%}   & \multicolumn{1}{c|}{0.3\%} & \multicolumn{1}{c|}{3\%}   & 100\% \\ \hline
\multicolumn{1}{|c|}{\multirow{2}{*}{\textbf{Zero-shot}}} & \multicolumn{1}{c|}{\multirow{2}{*}{CLIP-V}} & \multirow{2}{*}{CLIP-T} & IQA                   & \multicolumn{1}{c|}{50.27} & \multicolumn{1}{c|}{}      & \multicolumn{1}{c|}{}      & \multicolumn{1}{c|}{}      & \multicolumn{1}{c|}{58.3}  & \multicolumn{1}{c|}{}      & \multicolumn{1}{c|}{}      &       \\ \cline{4-12}
\multicolumn{1}{|c|}{}                                    & \multicolumn{1}{c|}{}                        &                         & IA                    & \multicolumn{1}{c|}{54.82} & \multicolumn{1}{c|}{}      & \multicolumn{1}{c|}{}      & \multicolumn{1}{c|}{}      & \multicolumn{1}{c|}{53.82} & \multicolumn{1}{c|}{}      & \multicolumn{1}{c|}{}      &       \\ \hline
\multicolumn{3}{|c|}{}                                                                                                             & \textbf{Adapter Head} & \multicolumn{8}{c|}{}                                                                                                                                                                                            \\ \hline
\multicolumn{1}{|c|}{\multirow{3}{*}{\textbf{Adapter}}}   & \multicolumn{1}{c|}{\multirow{3}{*}{CLIP-V}} & \multirow{3}{*}{CLIP-T} & 1 Linear              & \multicolumn{1}{c|}{}      & \multicolumn{1}{c|}{30.13} & \multicolumn{1}{c|}{30.39} & \multicolumn{1}{c|}{31.86} & \multicolumn{1}{c|}{}      & \multicolumn{1}{c|}{28.1}  & \multicolumn{1}{c|}{27.43} & 29.31 \\ \cline{4-12}
\multicolumn{1}{|c|}{}                                    & \multicolumn{1}{c|}{}                        &                         & 3 Linear              & \multicolumn{1}{c|}{}      & \multicolumn{1}{c|}{30.43} & \multicolumn{1}{c|}{30.86} & \multicolumn{1}{c|}{32.31} & \multicolumn{1}{c|}{}      & \multicolumn{1}{c|}{28.73} & \multicolumn{1}{c|}{28.32} & 29.94 \\ \cline{4-12}
\multicolumn{1}{|c|}{}                                    & \multicolumn{1}{c|}{}                        &                         & 1 Transformer         & \multicolumn{1}{c|}{}      & \multicolumn{1}{c|}{34.41} & \multicolumn{1}{c|}{36.34} & \multicolumn{1}{c|}{36.42} & \multicolumn{1}{c|}{}      & \multicolumn{1}{c|}{33.35} & \multicolumn{1}{c|}{34.02} & 35.48 \\ \hline
\end{tabular}%
}
\caption{Evaluation of zero-shot and adding adapters performance on CLIP-V, CLIP-T. IQA represents that we group both question and answer together as the text prompt. When calculating the cosine distance in each sample, we measure the distance between the image feature and the text prompt feature.}
\label{vcr_adapterl}
\end{table*}

 \begin{table*}[h!]
\centering
\resizebox{\textwidth}{!}{%
\begin{tabular}{|c|l|c|c|c|c|c|c|c|ccc|ccc|}
\hline
\textbf{Method}                                                                                   & \textbf{Variation}                                                                                             & \textbf{V.} & \textbf{L.} & \textbf{TS}                                          & \textbf{CW} & \textbf{AF} & \textbf{\begin{tabular}[c]{@{}c@{}}Distillation\\ Weight\end{tabular}}                                  & \textbf{Token \#} & \multicolumn{3}{c|}{\textbf{Standard   Evaluation}}                    & \multicolumn{3}{c|}{\textbf{SM Evaluation}}                            \\ \hline
                                                                                                  &                                                                                                                &             &             &                                                      &             &             &                                                                                                         &                   & \multicolumn{1}{c|}{0.3\%}    & \multicolumn{1}{c|}{3\%}       & 100\% & \multicolumn{1}{c|}{0.3\%}    & \multicolumn{1}{c|}{3\%}       & 100\% \\ \hline
                                                                                                  &                                                                                                                &             &             &                                                      &             &             &                                                                                                         &                   & \multicolumn{1}{c|}{100 SP/C} & \multicolumn{1}{c|}{1000 SP/C} &       & \multicolumn{1}{c|}{100 SP/C} & \multicolumn{1}{c|}{1000 SP/C} &       \\ \hline
\textbf{Baseline}                                                                                 &                                                                                                                & -           & -           & -                                                    & -           & -           &                                                                                                         &                   & \multicolumn{1}{c|}{30.85}    & \multicolumn{1}{c|}{53.48}     & 75.53 & \multicolumn{1}{c|}{26.37}    & \multicolumn{1}{c|}{49.27}     & 71.13 \\ \hline
\textbf{\begin{tabular}[c]{@{}c@{}}Unimodal\\ Distillation\end{tabular}}                          &                                                                                                                & Y           &             &                                                      &             &             & $0.05$                                                                                                  &                   & \multicolumn{1}{c|}{34.24}    & \multicolumn{1}{c|}{54.53}     & 75.92 & \multicolumn{1}{c|}{31.11}    & \multicolumn{1}{c|}{51.16}     & 72.24 \\ \hline
\multirow{5}{*}{\textbf{\begin{tabular}[c]{@{}c@{}}Multimodal \\ Distillation (MD $\star$)\end{tabular}}} & \multicolumn{1}{c|}{\multirow{5}{*}{\begin{tabular}[c]{@{}c@{}}Different \\ Distillation Weight\end{tabular}}} & Y           & Y           &                                                      &             &             & $1.0$                                                                                                   &                   & \multicolumn{1}{c|}{33.39}    & \multicolumn{1}{c|}{51.57}     & 75.85 & \multicolumn{1}{c|}{33.22}    & \multicolumn{1}{c|}{50.05}     & 71.65 \\ \cline{3-15}
                                                                                                  & \multicolumn{1}{c|}{}                                                                                          & Y           & Y           &                                                      &             &             & $0.5$                                                                                                   &                   & \multicolumn{1}{c|}{34.20}    & \multicolumn{1}{c|}{52.96}     & 75.89 & \multicolumn{1}{c|}{33.31}    & \multicolumn{1}{c|}{51.17}     & 72.10 \\ \cline{3-15}
                                                                                                  & \multicolumn{1}{c|}{}                                                                                          & Y           & Y           &                                                      &             &             & $0.1$                                                                                                   &                   & \multicolumn{1}{c|}{34.47}    & \multicolumn{1}{c|}{54.72}     & 76.10 & \multicolumn{1}{c|}{33.64}    & \multicolumn{1}{c|}{51.32}     & 73.20 \\ \cline{3-15}
                                                                                                  & \multicolumn{1}{c|}{}                                                                                          & Y           & Y           &                                                      &             &             & $0.05$                                                                                                  &                   & \multicolumn{1}{c|}{36.78}    & \multicolumn{1}{c|}{55.91}     & 76.37 & \multicolumn{1}{c|}{34.93}    & \multicolumn{1}{c|}{52.06}     & 73.29 \\ \cline{3-15}
                                                                                                  & \multicolumn{1}{c|}{}                                                                                          & Y           & Y           &                                                      &             &             & $0.01$                                                                                                  &                   & \multicolumn{1}{c|}{35.31}    & \multicolumn{1}{c|}{53.65}     & 76.23 & \multicolumn{1}{c|}{31.34}    & \multicolumn{1}{c|}{51.23}     & 72.03 \\ \hline
\multirow{4}{*}{\textbf{+ Token Selection}}                                                       & \multirow{3}{*}{Different Token \#}                                                                            & Y           & Y           & Y                                                    &             &             & \multirow{5}{*}{\begin{tabular}[c]{@{}c@{}}Adaptive\\ $0.05 \cdot \frac{Token \# + 2}{2}$\end{tabular}} & 1                 & \multicolumn{1}{c|}{36.59}    & \multicolumn{1}{c|}{56.49}     & 76.37 & \multicolumn{1}{c|}{34.70}    & \multicolumn{1}{c|}{52.03}     & 72.30 \\ \cline{3-7} \cline{9-15}
                                                                                                  &                                                                                                                & Y           & Y           & Y                                                    &             &             &                                                                                                         & 2                 & \multicolumn{1}{c|}{37.21}    & \multicolumn{1}{c|}{57.64}     & 76.83 & \multicolumn{1}{c|}{36.14}    & \multicolumn{1}{c|}{53.33}     & 73.78 \\ \cline{3-7} \cline{9-15}
                                                                                                  &                                                                                                                & Y           & Y           & Y                                                    &             &             &                                                                                                         & 3                 & \multicolumn{1}{c|}{35.25}    & \multicolumn{1}{c|}{54.04}     & 76.16 & \multicolumn{1}{c|}{35.66}    & \multicolumn{1}{c|}{51.56}     & 72.41 \\ \cline{2-7} \cline{9-15}
                                                                                                  & \begin{tabular}[c]{@{}l@{}}w/o language prior\\ (random selection)\end{tabular}                                & Y           & Y           & \begin{tabular}[c]{@{}c@{}}Y\\ (Random)\end{tabular} &             &             &                                                                                                         & 2                 & \multicolumn{1}{c|}{30.91}    & \multicolumn{1}{c|}{53.24}     & 75.34 & \multicolumn{1}{c|}{29.02}    & \multicolumn{1}{c|}{50.43}     & 72.13 \\ \cline{1-7} \cline{9-15}
\textbf{\begin{tabular}[c]{@{}c@{}}+ Confidence \\ Weighting (CW)\end{tabular}}                   &                                                                                                                & Y           & Y           & Y                                                    & Y           &             &                                                                                                         & 2                 & \multicolumn{1}{c|}{38.28}    & \multicolumn{1}{c|}{58.68}     & 77.15 & \multicolumn{1}{c|}{37.82}    & \multicolumn{1}{c|}{54.12}     & 74.24 \\ \hline
\multirow{2}{*}{\textbf{\begin{tabular}[c]{@{}c@{}}+ Adaptive\\ \\ Finetuning (AF)\end{tabular}}} & \begin{tabular}[c]{@{}l@{}}w/o distillation \\ (2nd-stage "pretraining")\end{tabular}                          & Y           & Y           & Y                                                    & Y           & Y           & -                                                                                                       &                   & \multicolumn{1}{c|}{38.32}    & \multicolumn{1}{c|}{54.13}     & 76.38 & \multicolumn{1}{c|}{36.42}    & \multicolumn{1}{c|}{53.07}     & 72.87 \\ \cline{2-15}
                                                                                                  & \begin{tabular}[c]{@{}l@{}}w/ distillation\\ (MAD $\star$)\end{tabular}                                                & Y           & Y           & Y                                                    & Y           & Y           & \multirow{2}{*}{\begin{tabular}[c]{@{}c@{}}Adaptive\\ $0.05 \cdot \frac{Token \# + 2}{2}$\end{tabular}} & 2                 & \multicolumn{1}{c|}{39.02}    & \multicolumn{1}{c|}{56.23}     & 76.93 & \multicolumn{1}{c|}{38.26}    & \multicolumn{1}{c|}{55.87}     & 74.04 \\ \cline{1-7} \cline{9-15}
\textbf{MAD (S)}                                                                                  & + Semi-supervision                                                                                             & Y           & Y           & Y                                                    & Y           & Y           &                                                                                                         & 2                 & \multicolumn{1}{c|}{54.43}    & \multicolumn{1}{c|}{61.33}     & 77.61 & \multicolumn{1}{c|}{46.71}    & \multicolumn{1}{c|}{58.18}     & 74.55 \\ \hline
 \vspace{20mm}
\end{tabular}%
}

\caption{VCR distillation ablation experiments using VL-BERT student model. Training data subsampling shown under evaluation protocol. SM = Shortcut Mitigated.}
\label{tab:vcrablation}
\end{table*}

%  \begin{table*}[h!]
% \centering
% \resizebox{4cm}{!}{%
% \begin{tabular}{|c|cc|cc|}
% \hline
% \textbf{Method}  & \multicolumn{2}{c|}{\textbf{Std. Evaluation}} & \multicolumn{2}{c|}{\textbf{SM   Evaluation}} \\ \hline
%                  & \multicolumn{1}{c|}{0.3 \%}      & 3 \%       & \multicolumn{1}{c|}{0.3 \%}      & 3 \%       \\ \hline
% -                & \multicolumn{1}{c|}{46.74}       & 59.98      & \multicolumn{1}{c|}{45.25}       & 55.86      \\ \hline
% \textbf{MD (S)}  & \multicolumn{1}{c|}{51.32}       & 60.21      & \multicolumn{1}{c|}{48.24}       & 57.53      \\ \hline
% \textbf{MAD (S)} & \multicolumn{1}{c|}{52.43}       & 61.33      & \multicolumn{1}{c|}{48.71}       & 58.82      \\ \hline
% \end{tabular}%
% }
% \end{table*}

\section{Evaluation}
\subsection{VCR}
\label{sec:vcr}
\subsubsection{Public Leaderboard Results}
\label{sec:vcr-publicresults}

We submitted our single model test prediction results to the VCR public leaderboard, which is currently listed as the 12th entry overall (including ensemble models and models without any reference or publication) and achieves State-Of-The-Art (\textbf{SOTA}) performance on VCR compared to other public single models \footnote{single models with public reference or publication} that are pretrained with image-text data. The entry is displayed under the name, \textbf{CLIP-TD}\footnote{A former name with CLIP's vision and text encoders as the teacher models.} with test result: Q2A $79.6 \%$ QA2R $82.9 \%$ Q2AR $66.2 \%$. For more thoroughly comparing our work with others, we also submitted its entry to the other VCR public leaderboard hosted by DARPA. It is ranked as the 4th overall entry with a comparatively weaker base model among our student models, VL-BERT \cite{Su2020VL-BERT}. Meanwhile, we are also submitting our ensemble models' prediction results to the leaderboard, as shown in the second last row of Tab. \ref{vcr_full}.

\subsubsection{Standard Validation Results}
\label{sec:vcr-valresults}

As in Tab. \ref{vcr_full}, we include comparison of our methods on top of top-performing base models for all three evaluation metrics: Q2A, QA2R and Q2AR. Our method consistently improves on top of the base models across all metrics.

\subsubsection{Low-shot}
For every low-shot experiment across VCR, VQA and SNLI-VE, we average every experiment's result over at least four runs. The corresponding variance is also listed in Tab. \ref{tab:vcrlowshot}, Tab. \ref{tab:snlivelowshot} and Tab. \ref{tab:vqalowshot}.

\begin{table}[h!]
\centering
\resizebox{\textwidth}{!}{%
\begin{tabular}{|c|c|cc|cccc|cccc|}
\hline
\textbf{\begin{tabular}[c]{@{}c@{}}Base (Student)\\ Model\end{tabular}} & \textbf{Method}     & \multicolumn{2}{c|}{\textbf{Teacher Model}}                            & \multicolumn{4}{c|}{\textbf{Standard Evaluation}}                                                                                  & \multicolumn{4}{c|}{SM   Evaluation}                                                                                                \\ \hline
\multirow{2}{*}{}                                                       &                     & \multicolumn{1}{c|}{VE}                      & TE                      & \multicolumn{1}{c|}{0.3\%}                  & \multicolumn{1}{c|}{Var} & \multicolumn{1}{c|}{3\%}                    & Var & \multicolumn{1}{c|}{0.3\%}                   & \multicolumn{1}{c|}{Var} & \multicolumn{1}{c|}{3\%}                    & Var \\ \cline{2-12}
                                                                        &                     & \multicolumn{1}{c|}{}                        &                         & \multicolumn{1}{c|}{100 SP/C}               & \multicolumn{1}{c|}{}        & \multicolumn{1}{c|}{1000 SP/C}              &         & \multicolumn{1}{c|}{100 SP/C}                & \multicolumn{1}{c|}{}        & \multicolumn{1}{c|}{1000 SP/C}              &         \\ \hline
\multirow{3}{*}{VL-BERT}                                                & \textbf{Baseline}   & \multicolumn{1}{c|}{-}                       & -                       & \multicolumn{1}{c|}{30.85}                  & \multicolumn{1}{c|}{1.04}    & \multicolumn{1}{c|}{53.48}                  & 0.28    & \multicolumn{1}{c|}{26.37}                   & \multicolumn{1}{c|}{0.47}    & \multicolumn{1}{c|}{49.27}                  & 0.83    \\ \cline{2-12}
                                                                        & \textbf{MD$\star$}  & \multicolumn{1}{c|}{\multirow{2}{*}{CLIP-V}} & \multirow{2}{*}{CLIP-T} & \multicolumn{1}{c|}{36.78}                  & \multicolumn{1}{c|}{1.48}    & \multicolumn{1}{c|}{55.91}                  & 0.85    & \multicolumn{1}{c|}{34.93}                   & \multicolumn{1}{c|}{0.85}    & \multicolumn{1}{c|}{52.06}                  & 1.35    \\ \cline{2-2} \cline{5-12}
                                                                        & \textbf{MAD$\star$} & \multicolumn{1}{c|}{}                        &                         & \multicolumn{1}{c|}{\textbf{40.43 (+9.58)$\blacklozenge$}} & \multicolumn{1}{c|}{0.44}    & \multicolumn{1}{c|}{\textbf{58.98 (+5.5)$\blacklozenge$}}  & 0.51    & \multicolumn{1}{c|}{\textbf{39.27 (+12.9)$\blacklozenge$}}  & \multicolumn{1}{c|}{0.28}    & \multicolumn{1}{c|}{\textbf{54.88 (+5.61)$\blacklozenge$}} & 0.61    \\ \hline
\multirow{3}{*}{VILLA}                                                  & \textbf{Baseline}   & \multicolumn{1}{c|}{-}                       & -                       & \multicolumn{1}{c|}{34.84}                  & \multicolumn{1}{c|}{0.87}    & \multicolumn{1}{c|}{57.01}                  & 0.72    & \multicolumn{1}{c|}{29.41}                   & \multicolumn{1}{c|}{0.90}    & \multicolumn{1}{c|}{54.15}                  & 1.34    \\ \cline{2-12}
                                                                        & \textbf{MAD$\star$} & \multicolumn{1}{c|}{\multirow{2}{*}{CLIP-V}} & CLIP-T                  & \multicolumn{1}{c|}{42.95}                  & \multicolumn{1}{c|}{1.14}    & \multicolumn{1}{c|}{60.93}                  & 1.51    & \multicolumn{1}{c|}{41.97}                   & \multicolumn{1}{c|}{0.59}    & \multicolumn{1}{c|}{55.20}                  & 1.07    \\ \cline{2-2} \cline{4-12}
                                                                        & \textbf{}           & \multicolumn{1}{c|}{}                        & RoBERTa                 & \multicolumn{1}{c|}{\textbf{43.11 (+8.27)$\blacklozenge$}} & \multicolumn{1}{c|}{1.32}    & \multicolumn{1}{c|}{\textbf{61.49 (+4.48)$\blacklozenge$}} & 0.89    & \multicolumn{1}{c|}{\textbf{41.98 (+12.57)$\blacklozenge$}} & \multicolumn{1}{c|}{1.15}    & \multicolumn{1}{c|}{\textbf{56.85 (+2.7)$\blacklozenge$}}  & 1.23    \\ \hline
\end{tabular}%
}
\caption{Low-shot evaluation on VCR with variance. Every low-shot experiment result is averaged over four runs. $\blacklozenge$ The value in () representing the difference against the baseline.}
\label{tab:vcrlowshot}
\end{table}

\subsubsection{Shortcut Mitigation Results}
\label{sec:vcr-sm-valresults}
\begin{figure}[th!]
% \vspace{-4mm}
\begin{center}
\scriptsize
\resizebox{12cm}{!}{%
 \includegraphics[]{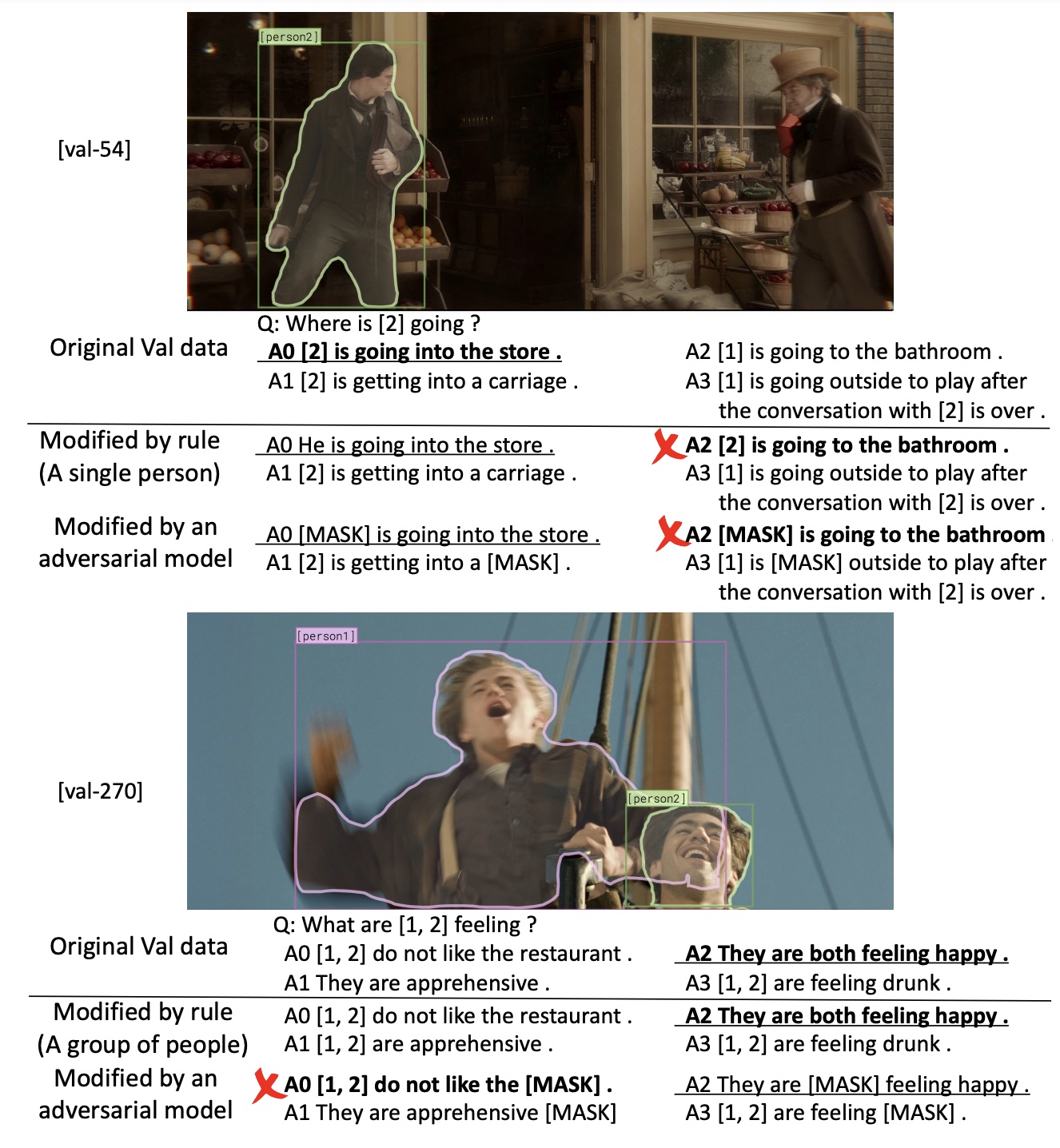}
}
\end{center}
% \vspace{-6mm}%Put here to reduce too much white space after your table
 \caption{Examples of modified samples in SM validation set.}
\label{fig:debias2}
% \vspace{-4mm}%Put here to reduce too much white space after your table
\end{figure}

For better evaluating our algorithms' performance in settings in real world scenarios where domain shift  may often occur. We follow \cite{debias}'s modification and compare our method against others in its modified validation set. The evaluation results are listed in Tab. 1 and Tab. 2 of the main paper submission. The modified example is shown in Fig. \ref{fig:debias2}. Motivated by the observation that "The correct option has the most overlap with the question" as showned in  Fig. \ref{fig:debias} and stated in \cite{debias}, the modification mainly focus on changing the pronouns of the correct and incorrect answer choices. As showned in Fig. \ref{fig:debias2}, the correct answers' pronouns are changed to be different from the question and the incorrect answer choices' pronouns are changed to be the same, instead.

\begin{figure}[th!]
% \vspace{-4mm}
\begin{center}
\scriptsize
\resizebox{12cm}{!}{%
 \includegraphics[]{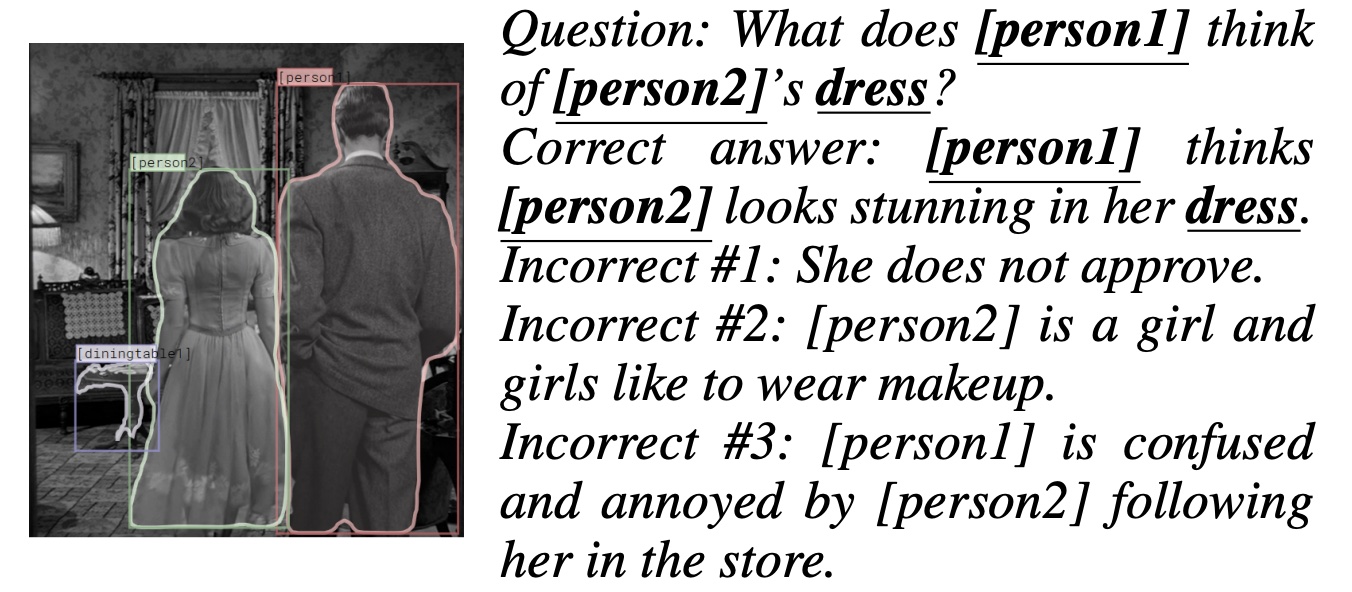}
}
\end{center}
% \vspace{-6mm}%Put here to reduce too much white space after your table
 \caption{An example of the standard validation set in VCR. The correct answer has the same pronoun as the question while the incorrect answer choices may not.}
\label{fig:debias1}
% \vspace{-4mm}%Put here to reduce too much white space after your table
\end{figure}

\subsubsection{VCR Dataset Question Sub-Types}
\label{sec:vcr-subtypes}

According to \cite{zellers2019vcr}, among VCR questions, 38\% fall into explanation (why, how come, \textit{etc.}), 24\% activity (doing, looking, event, \textit{etc.}), 13\% temporal (happened, before, after, \textit{etc.}), 8\% mental (feeling, thinking, \textit{etc.}), 7\% role (relations, occupations, \textit{etc.}), 5\% scene (where, near, \textit{etc.}), and 5\% hypothetical (if, would, could, \textit{etc.}). Details can be referred to \cite{zellers2019vcr}.

\subsubsection{Further Analysis of Multimodal Distillation}
\label{sec:vcr-analysis}

Based on our results, we have seen the impact on end task performance of distilling knowledge from largely pretrained unimodal encoders into student VL models. However, a key question still remains: how much the improvement comes from the distillation of each modality respectively?

Following \cite{cao2020behind}, we measure the Modality Importance (MI) of both visual modality and textual modality. This approach sums the attention weights across heads of each modality to understand how much each modality is weighted by the model. Fig. \ref{fig:mi_line} shows the average the MI values of all the heads for each layer on VL-BERT, both with and without MD, trained on the VCR dataset. One can clearly observe that prior to distillation, the model more heavily weights the text modality as being important to correctly choosing answers. After distillation, both vision and text modalities are more equally considered. This may also explain why the model yields such impressive performance improvements in low-shot and domain shifted scenarios.

Fig. \ref{fig:mi_line}, we plot the MI values of all the heads across 12 layers in VL-BERT Base and VL-BERT Base with MD. It is obvious that, at the last layer, the textual MI heatmap on the right is denser than the visual MI heatmap on the left. This shows a common flaw from existing V+L models that they heavily rely on the textual information than the visual part indicating the shallow understanding of the visual scene in downstream tasks. However, in the bottom row, the difference between the left and right heatmaps is much smaller and the visual MI heatmap at the bottom is also clearly more denser than the one at the top. This could be further verified by

% Compared to concurrent fine-tuning work CLIP-ViL, our approach outperforms under all low-shot conditions and most fully-supervised conditions, except for fully-supervised VQA, where the performance is comparable. Because our approach is agnostic to the base student model, ensembles of many student models can yield superior results under all conditions studied. We believe our results can be helpful for the community to know how to best leverage CLIP to potentially boost the performance of future works on VL tasks.

% \paragraph{Acknowledgement}
% Thanks to Liunian Harold Li for his help in the implementation of CLIP-ViL and feedbacks of the idea.

% \clearpage\mbox{}Page \thepage\ of the manuscript.
% \clearpage\mbox{}Page \thepage\ of the manuscript.

% This is the last page of the manuscript.
% Now we have reached the maximum size of the ECCV 2022 submission (excluding references).
% References should start immediately after the main text, but can continue on p.15 if needed.

% \clearpage
\subsection{VQA}
\subsubsection{Zero-shot}
VQA is different from VCR and SNLI-VE in terms of number of answer labels for each sample data. For every image-question pair, VCR provides four answer choices and for every premise-hypothesis pair, only three answer labels are provided. While in VQA, for every image-question pair, more than 2000 answer labels are provided. Thus the format of VQA does not strictly follow the conventional Multiple-Choice-Question (MCQ) format. Given the vast variety of answer labels provided, every question in VQA may potentially map to more than one answer choice. This one-to-many mappings may introduce difficulty in evaluating learning algorithms' ability of answering the VQA questions correctly. Different from \cite{shengshen} which naively measure the distance between image-question pair against all the answer labels and deliver a performance of almost 0 $\%$. We highly suspect that is the objective solution in terms of evaluating largely pretrained unimodal encoders' generalization in solving VQA tasks.

In this work, we first apply fixed number of templates embedded with heuristic knowledge and convert every question into statement. Then we further utilize pretrained sentence embedding \cite{reimers2019sentence} to measure the cosine distance between every question against all the answer labels. Based on the ranking of the cosine measurement, we filter all the answer labels to three set of answer candidates: Top 1, Top 3 and Top 10. We utilize CLIP's inference steps and prompt engineering and evaluate zero-shot performance of CLIP-V and CLIP-T across the three set, as shown in Tab. \ref{tab:vqazeroshot}.

\begin{table}[h!]
\centering
\resizebox{5cm}{!}{%
\begin{tabular}{|c|c|c|c|}
\hline
Question Type & Top 1 & Top 3 & Top 10 \\ \hline
Overall       & 21.77 & 53.67 & 71.64  \\ \hline
Binary        & 14.01 & 41.12 & 41.12  \\ \hline
Number        & 2.75  & 50.9  & 9.62   \\ \hline
Others        & 6.58  & 10.37 & 19.63  \\ \hline
\end{tabular}%
}
\caption{Zero-shot performance of CLIP-V and CLIP-T on VQA.}
\label{tab:vqazeroshot}
\end{table}

\subsubsection{Low-shot}
For every low-shot experiment across VCR, VQA and SNLI-VE, we average every experiment's result over at least four runs. The corresponding variance is also listed in Tab. \ref{tab:vcrlowshot}, Tab. \ref{tab:snlivelowshot} and Tab. \ref{tab:vqalowshot}.

\begin{table}[h!]
\centering
\resizebox{10cm}{!}{%
\begin{tabular}{|c|c|cc|clcl|}
\hline
\textbf{\begin{tabular}[c]{@{}c@{}}Base (Student)\\ Model\end{tabular}} & \textbf{Method}       & \multicolumn{2}{c|}{\textbf{Teacher Model}}                                      & \multicolumn{4}{c|}{\textbf{Validation}}                                                                                                                               \\ \hline
                                                                        &                       & \multicolumn{1}{c|}{\multirow{2}{*}{\textbf{VE}}} & \multirow{2}{*}{\textbf{TE}} & \multicolumn{2}{c|}{\begin{tabular}[c]{@{}c@{}}0.3 $\%$\\ (100 SP/C)\end{tabular}} & \multicolumn{2}{c|}{\begin{tabular}[c]{@{}c@{}}3 $\%$\\ (1000 SP/C)\end{tabular}} \\ \cline{1-2} \cline{5-8}
\multicolumn{1}{|l|}{}                                                  & \multicolumn{1}{l|}{} & \multicolumn{1}{c|}{}                             &                              & \multicolumn{1}{c|}{Accuracy}                     & \multicolumn{1}{l|}{Var}       & \multicolumn{1}{c|}{Accuracy}                               & Var                 \\ \hline
\multirow{3}{*}{VL-BERT}                                                & Baseline              & \multicolumn{1}{c|}{-}                            & -                            & \multicolumn{1}{c|}{53.28}                        & \multicolumn{1}{l|}{0.12}      & \multicolumn{1}{c|}{62.31}                                  & 0.46                \\ \cline{2-8}
                                                                        & \textbf{MD$\star$}    & \multicolumn{1}{c|}{\multirow{2}{*}{CLIP-V}}      & \multirow{2}{*}{CLIP-T}      & \multicolumn{1}{c|}{56.02}                        & \multicolumn{1}{l|}{0.46}      & \multicolumn{1}{c|}{64.92}                                  & 0.74                \\ \cline{2-2} \cline{5-8}
                                                                        & \textbf{MAD$\star$}   & \multicolumn{1}{c|}{}                             &                              & \multicolumn{1}{c|}{\textbf{56.78 (+3.5)$\blacklozenge$}}        & \multicolumn{1}{l|}{1.31}      & \multicolumn{1}{c|}{\textbf{65.37 (+3.06)$\blacklozenge$}}                 & 1.09                \\ \hline
\multirow{3}{*}{VILLA}                                                  & Baseline              & \multicolumn{1}{c|}{-}                            & -                            & \multicolumn{1}{c|}{58.47}                        & \multicolumn{1}{l|}{0.76}      & \multicolumn{1}{c|}{67.16}                                  & 0.28                \\ \cline{2-8}
                                                                        & \textbf{MAD$\star$}   & \multicolumn{1}{c|}{\multirow{2}{*}{CLIP-V}}      & CLIP-T                       & \multicolumn{1}{c|}{\textbf{59.65 (+1.18)$\blacklozenge$}}       & \multicolumn{1}{l|}{0.13}      & \multicolumn{1}{c|}{68.43}                                  & 0.15                \\ \cline{2-2} \cline{4-8}
                                                                        &                       & \multicolumn{1}{c|}{}                             & RoBERTa                      & \multicolumn{1}{c|}{58.48}                        & \multicolumn{1}{l|}{0.43}      & \multicolumn{1}{c|}{\textbf{68.86 (+1.70)$\blacklozenge$}}                 & 0.66                \\ \hline
\end{tabular}%
}
\caption{Low-shot evaluation on VQA with variance. Every low-shot experiment result is averaged over four runs. $\blacklozenge$ The value in () representing the difference against the baseline.}
\label{tab:vqalowshot}
\end{table}

\subsection{SNLI-VE}
\subsubsection{Low-shot}

For every low-shot experiment across VCR, VQA and SNLI-VE, we average every experiment's result over at least four runs. The corresponding variance is also listed in Tab. \ref{tab:vcrlowshot}, Tab. \ref{tab:snlivelowshot} and Tab. \ref{tab:vqalowshot}.

\begin{table}[h!]
\centering
\resizebox{10cm}{!}{%
\begin{tabular}{|c|c|cc|cccc|}
\hline
\textbf{\begin{tabular}[c]{@{}c@{}}Base (Student)\\ Model\end{tabular}} & \textbf{Method}     & \multicolumn{2}{c|}{\textbf{Teacher Model}}                            & \multicolumn{4}{c|}{\textbf{Validation}}                                                                                                                         \\ \hline
                                                                        &                     & \multicolumn{1}{c|}{\textbf{VE}}             & TE                      & \multicolumn{2}{c|}{\begin{tabular}[c]{@{}c@{}}0.3\%\\ (100 SP/C)\end{tabular}} & \multicolumn{2}{c|}{\begin{tabular}[c]{@{}c@{}}3\%\\ (1000 SP/C)\end{tabular}} \\ \hline
                                                                        &                     & \multicolumn{1}{c|}{}                        &                         & \multicolumn{1}{c|}{Accuracy}                   & \multicolumn{1}{c|}{Var}      & \multicolumn{1}{c|}{Accuracy}                             & Var                \\ \hline
\multirow{3}{*}{VL-BERT}                                                & Baseline            & \multicolumn{1}{c|}{-}                       & -                       & \multicolumn{1}{c|}{35.33}                      & \multicolumn{1}{c|}{0.31}     & \multicolumn{1}{c|}{63.29}                                & 0.52               \\ \cline{2-8}
                                                                        & \textbf{MD$\star$}  & \multicolumn{1}{c|}{\multirow{2}{*}{CLIP-V}} & \multirow{2}{*}{CLIP-T} & \multicolumn{1}{c|}{36.83}                      & \multicolumn{1}{c|}{0.38}     & \multicolumn{1}{c|}{64.43}                                & 0.27               \\ \cline{2-2} \cline{5-8}
                                                                        & \textbf{MAD$\star$} & \multicolumn{1}{c|}{}                        &                         & \multicolumn{1}{c|}{\textbf{37.12 (+1.79)$\blacklozenge$}}     & \multicolumn{1}{c|}{0.58}     & \multicolumn{1}{c|}{\textbf{65.71 (+2.42)$\blacklozenge$}}               & 1.02               \\ \hline
\multirow{3}{*}{VILLA}                                                  & Baseline            & \multicolumn{1}{c|}{-}                       & -                       & \multicolumn{1}{c|}{37.18}                      & \multicolumn{1}{c|}{0.44}     & \multicolumn{1}{c|}{65.75}                                & 0.30               \\ \cline{2-8}
                                                                        & \textbf{MAD$\star$} & \multicolumn{1}{c|}{\multirow{2}{*}{CLIP-V}} & CLIP-T                  & \multicolumn{1}{c|}{\textbf{40.16 (+2.98)$\blacklozenge$}}     & \multicolumn{1}{c|}{1.09}     & \multicolumn{1}{c|}{\textbf{66.93(+1.18)$\blacklozenge$}}               & 0.82               \\ \cline{2-2} \cline{4-8}
                                                                        &                     & \multicolumn{1}{c|}{}                        & RoBERTa                 & \multicolumn{1}{c|}{39.04}                      & \multicolumn{1}{c|}{0.53}     & \multicolumn{1}{c|}{66.23}                                & 0.62               \\ \hline
\end{tabular}%
}
\caption{Low-shot evaluation on SNLI-VE with variance. Every low-shot experiment result is averaged over four runs. $\blacklozenge$ The value in () representing the difference against the baseline.}
\label{tab:snlivelowshot}
\end{table}

\section{Training Details}
\label{sec:trainingdetails}
\subsection{Baselines}
\label{sec:trainingdetails-baselines}

All the baseline experiment results with models including VL-BERT, UNITER Base, UNITER Large, VILLA and CLIP-ViL$_{p}$ are based on code provided by the authors, which we modified to include our distillation methods. CLIP-ViL$_{p}$ did not originally evaluate on VCR in their pre-print. However, we evaluated it on the VCR dataset.

\subsection{Implementation Details}
\label{sec:trainingdetails-implementation}

\noindent\textbf{- Token Selectioin: } Our ablation experiments with TS distillation show that when the number of tokens selected is 2, the highest performance is obtained, shown in Tab. \ref{tab:vcrablation}.

\noindent\textbf{- Confidence Weighting:} With Confidence Weighted (CW) knowledge distillation method, during our experiment, we find out that the performance would achieve the optimal gain when the distillation is conducted across all four question-answer pairs instead of question-correct-answer pair only, as in Tab. \ref{tab:vcrablation}.

\noindent\textbf{- Adaptive Finetuning (AF) with Contrastive Knowledge:} Before the last-step finetuning for the target downstream tasks, we conduct Adaptive Finetuning to adapt the model to the downstream domain. During the Adaptive Finetuning, the model is trained on the full training set. The loss contains the construction loss from the same set of pretraining tasks like Masked Language Modeling (MLM), Image-Text Matching (ITM), \textit{etc.} as in \cite{chen2020uniter}. Additionally, we also include Naive Knowledge Distillation thus the final loss also includes the knowledge distillation loss besides the construction loss.

\noindent\textbf{- VL-BERT: } We train our model for 30 epochs with warm-up steps of 1000, SGD optimizer. Initial learning rate is $7.0e-5$ and  decays by 0.1 at the 14th, 18th and 26th epoch. The gradient accumulation steps is set to be 4 on 8 NVIDIA V100 GPUs (32GB VRAM). The total number of layers for VL-BERT is 24 VL-BERT$_{Large}$.

\noindent\textbf{- UNITER: }  Started with warm up steps of 800, the model is trained with total steps of 8000. With AdamW optimizer, the intial learning rate is set to be $6e-05$ with weight decay of 0.01 and batch size of 4000. The gradient accumulation steps is set to be 5 on 4 NVIDIA TITAN RTX GPUs (24GB VRAM).

\noindent\textbf{- VILLA: } Warm up steps is set to be 1000 and total training steps is 10000. The intial learning rate is $6e-05$ with weight decay of 0.01 and AdamW optimizer. The training batch size is 1250. The gradient accumulation steps is set to be 8 on 8 NVIDIA TITAN RTX GPUs (24GB VRAM).

\noindent\textbf{- CLIP-ViL$_{p}$: } The model is trained for 20 epochs with batch size of 24. The optimizer is AdamW with a peak learning rate of 5 x $10^{-5}$.

\section{Baseline Method}
\label{sec:baselinemethod}
\subsection{Multimodal Distillation}
\label{sec:baselinemethod-nkd}

% \begin{figure}[]
% \centering
% % \scalebox{0.8}{
% \includegraphics[width=1\linewidth]{figures/naive.pdf}
% % }
% \caption{The proposed Naive Knowledge Distillation method. For a given task with an established base model as the student. The l1 loss is calculated respectively for both the visual and text representation features between the teacher and the student.}
% \vspace{-5mm}
% \label{fig:naive_kd}
% \end{figure}

As illustrated in the top rows of Tab. \ref{tab:vcrablation}, we experiment with a wide spectrum of disllation weight and realize that the performance is best optimized when the weight is set to be 0.05.
\subsection{Adapters over CLIP}
\label{sec:baselinemethod-adapters}
For more comprehensively exploring different options to utilize the pretrained unimodal encoders for downstream Vision-Language tasks, we also experimented to directly add adapters on top of unimodal encoders. As listed in Tab. \ref{vcr_adapterl}, we eseentially have experimented with adding either MLP or attention layers on top of CLIP-V and CLIP-T. However, due to the large gap between the size of finetuning data and CLIP's original pretraining data, the adapters' limited capacity fail to adapt the preatrained unimodel encoders for the downstream tasks efficiently.

\end{document}